\documentclass{article}

\usepackage{bm}
\usepackage{amsmath}
\usepackage{array}
\usepackage{amssymb}
\usepackage{makecell}

\usepackage{hyperref}
\usepackage{microtype}
\usepackage{graphicx}
\usepackage{subcaption}
\usepackage{arydshln}
\usepackage{booktabs}       % professional-quality tables
\usepackage{subfloat}
\usepackage{caption} % Needed for footnotes in captions

\usepackage[accepted]{icml2019}

\newcommand{\captionabove}[2][]{%
    \vskip-\abovecaptionskip
    \vskip+\belowcaptionskip
    \ifx\@nnil#1\@nnil
        \caption{#2}%
    \else
        \caption[#1]{#2}%
    \fi
    \vskip-\belowcaptionskip
}

\usepackage{tikz}
\usetikzlibrary{bayesnet}

\tikzstyle{invert} = [diamond, fill=white, draw=black, minimum size=10pt, inner sep=0pt, node distance=0.58]
\tikzstyle{invertsm} = [diamond, fill=white, draw=black, minimum size=10pt, inner sep=0pt, node distance=0.4]
\tikzstyle{vecinvert} = [diamond, fill=white, draw=black, minimum size=10pt, inner sep=0pt, node distance=0.58, line width=1pt]
\tikzstyle{vecinvertsm} = [diamond, fill=white, draw=black, minimum size=10pt, inner sep=0pt, node distance=0.4, line width=1pt]
\tikzstyle{constdot} = [circle, fill=black, inner sep=0pt, minimum size=6pt]

\newcommand{\invertedge}[4][]{ %
  \foreach \f in {#3} { %
    \foreach \x in {#2} { %
      \path (\x) edge[-,#1] (\f) ; %
    } ;
    \foreach \y in {#4} { %
      \path (\f) edge[-,#1] (\y) ; %
    } ;
  } ;
}

\newcommand{\invert}[5][]{ %
  \node[invert, label={[name=#2-caption]#3}, name=#2, #1,
  alias=#2-alias] {} ; %
  \invertedge {#4} {#2-alias} {#5} ; %
}
\newcommand{\invertsm}[5][]{ %
  \node[invertsm, label={[name=#2-caption]#3}, name=#2, #1,
  alias=#2-alias] {} ; %
  \invertedge {#4} {#2-alias} {#5} ; %
}
\newcommand{\vecinvert}[5][]{ %
  \node[vecinvert, label={[name=#2-caption]#3}, name=#2, #1,
  alias=#2-alias] {} ; %
  \invertedge {#4} {#2-alias} {#5} ; %
}
\newcommand{\vecinvertsm}[5][]{ %
  \node[vecinvertsm, label={[name=#2-caption]#3}, name=#2, #1,
  alias=#2-alias] {} ; %
  \invertedge {#4} {#2-alias} {#5} ; %
}

\usepackage [autostyle, english = american]{csquotes}
\MakeOuterQuote{"}

\begin{document}

\twocolumn[
\icmltitle{Latent Normalizing Flows for Discrete Sequences}
\begin{icmlauthorlist}
\icmlauthor{Zachary M.~Ziegler}{harvard}
\icmlauthor{Alexander M.~Rush}{harvard}
\end{icmlauthorlist}
\icmlaffiliation{harvard}{School of Engineering and Applied Sciences, Harvard University, Cambridge, MA, USA}
\icmlcorrespondingauthor{Zachary M.~Ziegler}{zziegler@g.harvard.edu}

\icmlkeywords{Normalizing flow, VAE, NLP, language-modeling}

\vskip 0.3in
]

\printAffiliationsAndNotice{}

\begin{abstract}
Normalizing flows are a powerful class of generative models for continuous random variables, showing both strong model flexibility and the potential for non-autoregressive generation. These benefits are also desired when modeling discrete random variables such as text, but directly applying normalizing flows to discrete sequences poses significant additional challenges. We propose a VAE-based generative model which jointly learns a normalizing flow-based distribution in the latent space and a stochastic mapping to an observed discrete space. In this setting, we find that it is crucial for the flow-based distribution to be highly multimodal. To capture this property, we propose several normalizing flow architectures to maximize model flexibility. Experiments consider common discrete sequence tasks of character-level language modeling and polyphonic music generation. Our results indicate that an autoregressive flow-based model can match the performance of a comparable autoregressive baseline, and a non-autoregressive flow-based model can improve generation speed with a penalty to performance.
\end{abstract}

\section{Introduction}

In the past several years, deep generative models have been shown to give strong modeling performance across a range of tasks, including autoregressive models \cite{Bahdanau2015}, latent variable models \cite{Kingma2014}, implicit generative models \cite{NIPS2014_5423}, and exact-likelihood models based on normalizing flows \cite{Dinh2016}. These methods provide different tradeoffs along the axes of sample quality, generation speed, availability of comparative metrics, and interpretability.

Generative models based on normalizing flows have seen recent success in problems involving continuous random variables \cite{Jul}. Normalizing flows represent the joint distribution of a high-dimensional continuous random variable via an invertible deterministic transformation from a base density \cite{Rezende2015,Kingma2016}. Flows have been explored both to increase the flexibility of the variational posterior in the context of variational autoencoders (VAEs), and directly as a generative model which is the focus of this work. Within the class of exact likelihood models, normalizing flows provide two key advantages: model flexibility and generation speed. Flows generalize continuous autoregressive models \cite{Papamakarios2017} and give more distributional flexibility. Furthermore, normalizing flows can be designed that are non-autoregressive during sampling \cite{Oord2017,Jul}, enabling fast parallel generation.

Unfortunately, due to their reliance on parameterized applications of the change-of-variables formula, it is challenging to directly apply normalizing flows to discrete random variables. Applying related methods, e.g. via discrete change of variables or a relaxation, is an important line of research but leads to significant additional challenges.

We study normalizing flows in the discrete setting with a \textit{latent normalizing flow} within the VAE framework, where the flow models a continuous representation of the discrete data via the prior. Previous works have found that using VAEs with discrete data often leads to posterior collapse, where the powerful likelihood model ignores the latent code when learning to model the data \cite{Bowman2015,Yang2017a}. To ensure that the continuous flow fully captures the dynamics of the discrete space we use a simple inputless emission model for the likelihood. Additionally, this is important to ensure that the potential speed benefits of the flow are not bottle-necked by an autoregressive likelihood.

%The prior distribution is implemented with a normalizing flow, while the posterior is a simple inputless emission model. The weakness of the posterior allows the flow to model all of the interesting and important interactions across time, giving access to the benefits of flows (flexibility and parallelizability) decoupled from the final discrete selection in the emission model. Compared to models which require an autoregressive factorization in the data space, our decomposition allows for non-autoregressive generation, a strong latent representation of the data, and a nearly fully differentiable generative process (up to the emission). The latter can be especially useful for e.g. GANs, which have traditionally encountered challenges in their application to discrete data such as text.

The key to the success of such a model is the ability of the flow to capture the high degree of multimodality typically found in discrete data, such that strong performance can be achieved without an autoregressive likelihood. The flexibility of normalizing flows allows us to design specific flow architectures which target this desiderata. To this end, we propose three normalizing flow architectures designed to maximize flexibility in order to capture multimodal discrete dynamics in a continuous space.

Experiments consider discrete latent generative models for character-level language modeling and polyphonic music modeling. We find that the latent flow model is able to describe the character-level dataset as well as a discrete autoregressive LSTM-based model, and is able to describe the polyphonic music datasets comparably to other autoregressive latent-variable models. We further find that the parallel-generation version of the model is able to generate sentences faster than the baseline model, with a penalty to modeling performance. Finally, we analyze the functionality of the model and demonstrate how it induces the high degree of multimodality needed to map between continuous and discrete spaces. Code is available at \url{https://github.com/harvardnlp/TextFlow}.

\section{Related Work}

% \paragraph{Nonlinear invertible functions}
% The class of normalizing flows studied in this work have previously been based on a scalar affine transformation. \citet{Huang2018} show that an invertible transformation function can be constructed that more closely resembles a neural network. They demonstrate that the added flexibility improves performance over the affine transformation on density estimation benchmarks. Their model introduces additional optimization challenges, however, and is not analytically invertible. In this work we similarly desire a more flexible transformation but require analytical invertibility.

\paragraph{Latent Variable Models for Sequences}
In the context of language modeling, \citet{Bowman2015} experiment with a VAE of fixed size continuous latent space and an autoregressive RNN decoder. In practice, the VAE encodes little information about the sentence in the latent space because the decoder is powerful enough to model the data well, reducing the VAE to a standard autoregressive model. Recent work has focused on increasing the amount of information the model places in the latent space, either by modifying the prior density \cite{Xu2018}, the decoder structure \cite{Yang2017a}, or the variational inference procedure \cite{Kim2018}, though in all cases the model still relies heavily on the discrete decoder. Our proposed model removes the discrete autoregressive decoder entirely.

Other methods construct VAEs for sequence data with a variable size latent variable composed of one latent vector per input token  \cite{Bayer2014,Chung2015, Gu2015}. While similar to the model proposed in this work in the layout of the latent dimensions, these models also include an autoregressive discrete decoder. \citet{NIPS2018_8004} consider a similar variation to our proposed approach, without an autoregressive decoder, and experiment with small-scale models and baselines.

To the best of our knowledge, no previous works explore the latent sequence model regime of an inputless discrete decoder while achieving even moderate performance compared to strong autoregressive baselines.

\paragraph{Non-Autoregressive Generation}
In the domain of natural images, \citet{Dinh2016} and \citet{Jul} propose flow-based models for non-autoregressive generation. Compared to state-of-the-art autoregressive models, their model performs both training and generation in parallel but suffers a penalty to model accuracy. In the domain of raw audio waveforms, \cite{Oord2017} demonstrate parity between a non-autoregressive flow model and an autoregressive model, achieved via distillation and additional task-specific losses.

In the domain of text \citet{Gu2017} propose a model which uses fertility scores as a latent variable, approaching the performance of autoregressive models. While this works for translation due to the aligned nature of the sentences, the fertility framework and required pre-trained autoregressive model preclude the technique from more general application. \citet{Lee2018} propose a deterministic model based on a denoising process to iteratively improve the quality of a non-autoregresively generated sentence. The authors demonstrate strong performance at neural machine translation, but the technique does not model the full distribution and requires a task-specific predetermined denoising process.

In an alternative approach for faster neural machine translation, \citet{Kaiser2018} propose to use a discrete latent space of variable but reduced size (e.g. 8x fewer tokens than the length of the sentence). While this technique speeds up the translation process, it remains serial. Furthermore, the method makes no claims about fully modeling the distribution of the data.

\section{Background: Normalizing Flows}
\label{sec:background}

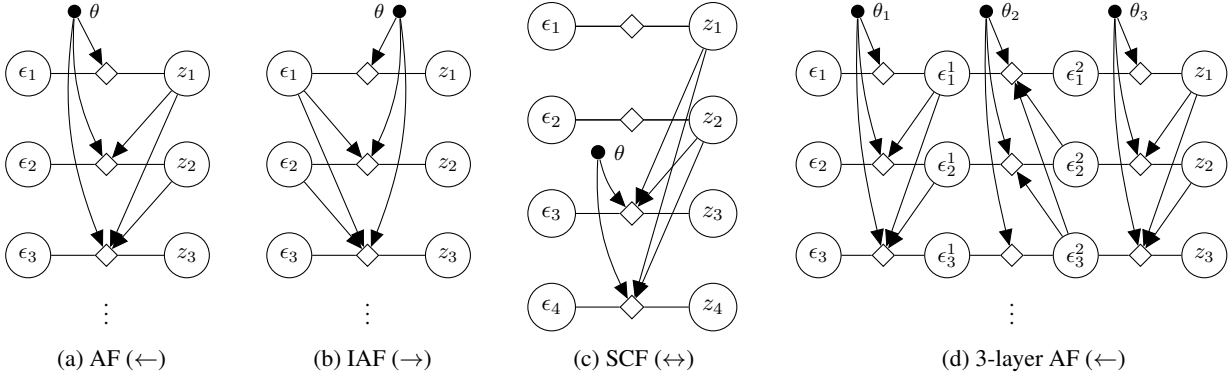
\begin{figure*}[t]
\centering
\begin{subfigure}[t]{.2\linewidth}
\hspace{.06\linewidth}
\begin{tikzpicture}[scale=0.85, every node/.style={scale=0.85}]
 % Nodes
 \node[latent]                   (eps1)      {$\epsilon_1$} ; %
 \node[latent, right=1.5cm of eps1]    (z1)      {$z_1$} ;
 \node[latent, below=0.6cm of eps1]    (eps2)      {$\epsilon_2$} ; %
 \node[latent, right=1.5cm of eps2]    (z2)      {$z_2$} ;
 \node[latent, below=0.6cm of eps2]    (eps3)      {$\epsilon_3$} ; %
 \node[latent, right=1.5cm of eps3]    (z3)      {$z_3$} ;
 %\node[latent, below=0.6cm of eps3]    (eps4)      {$\epsilon_4$} ; %
 %\node[latent, right=1.5cm of eps4]    (z4)      {$z_4$} ;

 % Inverts
 \invert[right=of eps1]     {eps-z1}     {} {eps1} {z1} ; %
 \invert[right=of eps2]     {eps-z2}     {} {eps2} {z2} ; %
 \invert[right=of eps3]     {eps-z3}     {} {eps3} {z3} ; %
 %\invert[right=of eps4]     {eps-z4}     {} {eps4} {z4} ; %

 % Edges
 \path (z1) edge [->, >={triangle 45}] (eps-z2) ;
 \path (z1) edge [->, >={triangle 45}] (eps-z3) ;
 %\path (z1) edge [->, >={triangle 45}] (eps-z4) ;
 \path (z2) edge [->, >={triangle 45}] (eps-z3) ;
 %\path (z2) edge [->, >={triangle 45}] (eps-z4) ;
 %\path (z3) edge [->, >={triangle 45}] (eps-z4) ;

 % Parameters
 \node[constdot, above=of eps-z1, xshift=-0.5cm, yshift=-0.5cm, label=right:$\theta$]     (params)   {} ;
 \path (params) edge [->, >={triangle 45}] (eps-z1) ;
 \path (params) edge [bend right=15, ->, >={triangle 45}] (eps-z2) ;
 \path (params) edge [bend right=15, ->, >={triangle 45}] (eps-z3) ;
 %\path (params) edge [bend right=15, ->, >={triangle 45}] (eps-z4) ;

 % Ellipsis
 \node[below=0.2cm of eps-z3]    (ellipsis)      {$\vdots$} ;

\end{tikzpicture}
\caption{AF ($\leftarrow$)}
\end{subfigure}\hfill
\begin{subfigure}[t]{.2\linewidth}
\hspace{.06\linewidth}
\begin{tikzpicture}[scale=0.85, every node/.style={scale=0.85}]
 % Nodes
 \node[latent]                   (eps1)      {$\epsilon_1$} ; %
 \node[latent, right=1.5cm of eps1]    (z1)      {$z_1$} ;
 \node[latent, below=0.6cm of eps1]    (eps2)      {$\epsilon_2$} ; %
 \node[latent, right=1.5cm of eps2]    (z2)      {$z_2$} ;
 \node[latent, below=0.6cm of eps2]    (eps3)      {$\epsilon_3$} ; %
 \node[latent, right=1.5cm of eps3]    (z3)      {$z_3$} ;
 %\node[latent, below=0.6cm of eps3]    (eps4)      {$\epsilon_4$} ; %
 %\node[latent, right=1.5cm of eps4]    (z4)      {$z_4$} ;

 % Inverts
 \invert[right=of eps1]     {eps-z1}     {} {eps1} {z1} ; %
 \invert[right=of eps2]     {eps-z2}     {} {eps2} {z2} ; %
 \invert[right=of eps3]     {eps-z3}     {} {eps3} {z3} ; %
 %\invert[right=of eps4]     {eps-z4}     {} {eps4} {z4} ; %

 % Edges
 \path (eps1) edge [->, >={triangle 45}] (eps-z2) ;
 \path (eps1) edge [->, >={triangle 45}] (eps-z3) ;
 %\path (eps1) edge [->, >={triangle 45}] (eps-z4) ;
 \path (eps2) edge [->, >={triangle 45}] (eps-z3) ;
 %\path (eps2) edge [->, >={triangle 45}] (eps-z4) ;
 %\path (eps3) edge [->, >={triangle 45}] (eps-z4) ;

 % Parameters
 \node[constdot, above=of eps-z1, xshift=0.5cm, yshift=-0.5cm, label=left:$\theta$]     (params)   {} ;
 \path (params) edge [->, >={triangle 45}] (eps-z1) ;
 \path (params) edge [bend left=15, ->, >={triangle 45}] (eps-z2) ;
 \path (params) edge [bend left=15, ->, >={triangle 45}] (eps-z3) ;
 %\path (params) edge [bend left=15, ->, >={triangle 45}] (eps-z4) ;

 % Ellipsis
 \node[below=0.2cm of eps-z3]    (ellipsis)      {$\vdots$} ;

\end{tikzpicture}
\caption{IAF ($\rightarrow$)}
\end{subfigure}\hfill
\begin{subfigure}[t]{.2\linewidth}
\hspace{.06\linewidth}
\begin{tikzpicture}[scale=0.85, every node/.style={scale=0.9}][scale=0.85, every node/.style={scale=0.9}]
 % Nodes
 \node[latent]                   (eps1)      {$\epsilon_1$} ; %
 \node[latent, right=1.5cm of eps1]    (z1)      {$z_1$} ;
 \node[latent, below=0.6cm of eps1]    (eps2)      {$\epsilon_2$} ; %
 \node[latent, right=1.5cm of eps2]    (z2)      {$z_2$} ;
 \node[latent, below=0.6cm of eps2]    (eps3)      {$\epsilon_3$} ; %
 \node[latent, right=1.5cm of eps3]    (z3)      {$z_3$} ;
 \node[latent, below=0.6cm of eps3]    (eps4)      {$\epsilon_4$} ; %
 \node[latent, right=1.5cm of eps4]    (z4)      {$z_4$} ;

 % Inverts
 \path (eps1) edge (z1) ;
 \path (eps2) edge (z2) ;
 \invert[right=of eps1]     {eps-z1}     {} {eps1} {z1} ; %
 \invert[right=of eps2]     {eps-z2}     {} {eps2} {z2} ; %
 \invert[right=of eps3]     {eps-z3}     {} {eps3} {z3} ; %
 \invert[right=of eps4]     {eps-z4}     {} {eps4} {z4} ; %

 % Edges
 \path (z1) edge [->, >={triangle 45}] (eps-z3) ;
 \path (z1) edge [->, >={triangle 45}] (eps-z4) ;
 \path (z2) edge [->, >={triangle 45}] (eps-z3) ;
 \path (z2) edge [->, >={triangle 45}] (eps-z4) ;

 % Parameters
 \node[constdot, above=of eps-z3, xshift=-0.5cm, yshift=-0.5cm, label=right:$\theta$]     (params)   {} ;
 \path (params) edge [bend right=15, ->, >={triangle 45}] (eps-z3) ;
 \path (params) edge [bend right=15, ->, >={triangle 45}] (eps-z4) ;

 % Ellipsis
% \node[below=0.2cm of eps-z4]    (ellipsis)      {$\vdots$} ;

\end{tikzpicture}
\caption{SCF ($\leftrightarrow$)}
\end{subfigure}\hspace{0.02\linewidth}
\begin{subfigure}[t]{.37\linewidth}
\begin{tikzpicture}[scale=0.85, every node/.style={scale=0.85}]
 % Nodes
 \node[latent]                   (eps1)      {$\epsilon_1$} ; %
 \node[latent, below=0.6cm of eps1]    (eps2)      {$\epsilon_2$} ; %
 \node[latent, below=0.6cm of eps2]    (eps3)      {$\epsilon_3$} ; %
 %\node[latent, below=0.6cm of eps3]    (eps4)      {$\epsilon_4$} ; %
 \node[latent, right=1.1cm of eps1]    (z2prime1)      {$\epsilon^{1}_1$} ;
 \node[latent, right=1.1cm of eps2]    (z2prime2)      {$\epsilon^1_2$} ;
 \node[latent, right=1.1cm of eps3]    (z2prime3)      {$\epsilon^1_3$} ;
% \node[latent, right=1.1cm of eps4]    (z2prime4)      {$z^1_4$} ;
 \node[latent, right=1.1cm of z2prime1]    (zprime1)      {$\epsilon^2_1$} ;
 \node[latent, right=1.1cm of z2prime2]    (zprime2)      {$\epsilon^2_2$} ;
 \node[latent, right=1.1cm of z2prime3]    (zprime3)      {$\epsilon^2_3$} ;
 %\node[latent, right=1.1cm of z2prime4]    (zprime4)      {$z'_4$} ;
 \node[latent, right=1.1cm of zprime1, align=right]    (z1)      {$z_1$} ;
 \node[latent, right=1.1cm of zprime2]    (z2)      {$z_2$} ;
 \node[latent, right=1.1cm of zprime3]    (z3)      {$z_3$} ;
 %\node[latent, right=1.1cm of zprime4]    (z4)      {$z_4$} ;

 % Inverts
 \invertsm[right=of eps1]     {eps-z2prime1}     {} {eps1} {z2prime1} ; %
 \invertsm[right=of eps2]     {eps-z2prime2}     {} {eps2} {z2prime2} ; %
 \invertsm[right=of eps3]     {eps-z2prime3}     {} {eps3} {z2prime3} ; %
 %\invertsm[right=of eps4]     {eps-z2prime4}     {} {eps4} {z2prime4} ; %
 \invertsm[right=of z2prime1]  {z2prime-zprime1}     {} {z2prime1} {zprime1} ; %
 \invertsm[right=of z2prime2]  {z2prime-zprime2}     {} {z2prime2} {zprime2} ; %
 \invertsm[right=of z2prime3]  {z2prime-zprime3}     {} {z2prime3} {zprime3} ; %
 %\invertsm[right=of z2prime4]  {z2prime-zprime4}     {} {z2prime4} {zprime4} ; %
 \invertsm[right=of zprime1]     {zprime-z1}     {} {zprime1} {z1} ; %
 \invertsm[right=of zprime2]     {zprime-z2}     {} {zprime2} {z2} ; %
 \invertsm[right=of zprime3]     {zprime-z3}     {} {zprime3} {z3} ; %
 %\invertsm[right=of zprime4]     {zprime-z4}     {} {zprime4} {z4} ; %

 % Edges
 \path (z2prime1) edge [->, >={triangle 45}] (eps-z2prime2) ;
 \path (z2prime1) edge [->, >={triangle 45}] (eps-z2prime3) ;
 %\path (z2prime1) edge [->, >={triangle 45}] (eps-z2prime4) ;
 \path (z2prime2) edge [->, >={triangle 45}] (eps-z2prime3) ;
 %\path (z2prime2) edge [->, >={triangle 45}] (eps-z2prime4) ;
 %\path (z2prime3) edge [->, >={triangle 45}] (eps-z2prime4) ;
 %\path (zprime4) edge [->, >={triangle 45}] (z2prime-zprime1) ;
 %\path (zprime4) edge [->, >={triangle 45}] (z2prime-zprime2) ;
 %\path (zprime4) edge [->, >={triangle 45}] (z2prime-zprime3) ;
 \path (zprime3) edge [->, >={triangle 45}] (z2prime-zprime1) ;
 \path (zprime3) edge [->, >={triangle 45}] (z2prime-zprime2) ;
 \path (zprime2) edge [->, >={triangle 45}] (z2prime-zprime1) ;
 \path (z1) edge [->, >={triangle 45}] (zprime-z2) ;
 \path (z1) edge [->, >={triangle 45}] (zprime-z3) ;
 %\path (z1) edge [->, >={triangle 45}] (zprime-z4) ;
 \path (z2) edge [->, >={triangle 45}] (zprime-z3) ;
 %\path (z2) edge [->, >={triangle 45}] (zprime-z4) ;
 %\path (z3) edge [->, >={triangle 45}] (zprime-z4) ;

 % Parameters
 \node[constdot, above=of eps-z2prime1, xshift=-0.4cm, yshift=-0.5cm, label=right:$\theta_1$]     (params1)   {} ;
 \path (params1) edge [->, >={triangle 45}] (eps-z2prime1) ;
 \path (params1) edge [bend right=8, ->, >={triangle 45}] (eps-z2prime2) ;
 \path (params1) edge [bend right=8, ->, >={triangle 45}] (eps-z2prime3) ;
 %\path (params1) edge [bend right=10, ->, >={triangle 45}] (eps-z2prime4) ;
 \node[constdot, above=of z2prime-zprime1, xshift=-0.4cm, yshift=-0.5cm, label=right:$\theta_2$]     (params2)   {} ;
 \path (params2) edge [->, >={triangle 45}] (z2prime-zprime1) ;
 \path (params2) edge [bend right=8, ->, >={triangle 45}] (z2prime-zprime2) ;
 \path (params2) edge [bend right=8, ->, >={triangle 45}] (z2prime-zprime3) ;
 %\path (params2) edge [bend right=10, ->, >={triangle 45}] (z2prime-zprime4) ;
 \node[constdot, above=of zprime-z1, xshift=-0.4cm, yshift=-0.5cm, label=right:$\theta_3$]     (params3)   {} ;
 \path (params3) edge [->, >={triangle 45}] (zprime-z1) ;
 \path (params3) edge [bend right=8, ->, >={triangle 45}] (zprime-z2) ;
 \path (params3) edge [bend right=8, ->, >={triangle 45}] (zprime-z3) ;
 %\path (params3) edge [bend right=10, ->, >={triangle 45}] (zprime-z4) ;

 % Ellipsis
 \node[below=0.2cm of z2prime-zprime3]    (ellipsis)      {$\vdots$} ;

\end{tikzpicture}
\caption{3-layer AF ($\leftarrow$)}
\end{subfigure}
\caption{\label{fig:standard_flows}Flow diagrams for normalizing flows acting on sequences of scalars. Circles represent random variables $\epsilon_d$ or $z_d$. Diamonds represent a parameterized invertible scalar transformation, $f_\theta$, in this case an affine transformation. Diagrams show the sampling process ($\boldsymbol{\epsilon} \rightarrow \boldsymbol{z}$, read left to right) and density evaluation ( $\boldsymbol{\epsilon} \leftarrow  \boldsymbol{z}$, read right to left). While all models can be used in both directions, they differ in terms of whether the calculation is serial or parallel, i.e. AF is parallel in evaluation but serial in sampling ($\leftarrow$) because $z_1$ is needed to sample $z_2$, whereas SCF is parallel for both ($\leftrightarrow$). }
\end{figure*}

Normalizing flows are a class of models that define a density through a parameterized invertible deterministic transformation from a base density, such as a standard Gaussian \cite{Tabak2010}. Define an invertible transformation $f_\theta : \mathcal{\epsilon} \rightarrow \mathcal{Z}$  and base density $p_\epsilon(\boldsymbol{\epsilon})$. These specify density $p_Z(\boldsymbol{z})$ via the change-of-variables formula:
\begin{equation*}
p_Z(\boldsymbol{z}) = p_\epsilon(f_\theta^{-1}(\boldsymbol{z})) \left|\text{det}\frac{\partial f_\theta^{-1}(\boldsymbol{z})}{\partial \boldsymbol{z}}\right|
\end{equation*}
Consider two core operations defined with flows:
%In particular we will be interested in two operations: sampling $z \sim p_Z$ and likelihood $p_Z(z)$ for a known $z$.
(a) Sampling, $\boldsymbol{z} \sim p_Z$, is performed by first sampling from the base distribution, $\boldsymbol{\epsilon} \sim p_{\epsilon}$, and then applying the forward transformation $\boldsymbol{z}=f_\theta(\boldsymbol{\epsilon})$;
(b) density evaluation, $p_Z(\boldsymbol{z})$ for a known $\boldsymbol{z}$, is computed by inverting the transformation, $\boldsymbol{\epsilon}=f_\theta^{-1}(\boldsymbol{z})$, and computing the base density $p_\epsilon(\boldsymbol{\epsilon})$. If $f_\theta$ is chosen to have an easily computable Jacobian determinant and inverse, both of these can be computed efficiently.

%Normalizing flow-based generative models operate in two directions: at training time observed variables $\boldsymbol{z}$ are given and $\boldsymbol{\epsilon}=f_\theta^{-1}(\boldsymbol{z})$ is used to evaluate (and then maximize) the likelihood of $\boldsymbol{z}$. At generation time $\boldsymbol{\epsilon}$ is sampled and $\boldsymbol{z}=f_\theta(\boldsymbol{\epsilon})$ is used to generate $\boldsymbol{z}$.

%Here we will focus on a class of normalizing flows, in the context of generative modeling, which construct dependencies between
One method for satisfying these criteria is to design functions that ensure a triangular Jacobian matrix and therefore a linear determinant calculation. We consider three common variants on this theme below. For this section we assume without loss of generality that $\mathcal{Z} = \mathbb{R}^D$ with ordered dimensions $1, \ldots, D$.

%In the summary below, we discuss three such flows addressing the common case where $\boldsymbol{z}$ is an observed random variable of size $T$.

\paragraph{Autoregressive Flow (AF)}
Autoregressive flows, originally proposed in \citet{Papamakarios2017}, ensure an invertible transformation and triangular Jacobian matrix by conditioning each scalar affine transformation on all previously observed variables $\boldsymbol{z}_{<d}$,
\begin{equation*}
\begin{gathered}
\label{eq:maf}
f_\theta(\boldsymbol{\epsilon})_d = z_d = a(\boldsymbol{z}_{<d}; \theta) + b(\boldsymbol{z}_{<d}; \theta) \cdot \epsilon_d \\
f_\theta^{-1}(\boldsymbol{z})_d = \epsilon_d = \frac{z_d - a(\boldsymbol{z}_{<d}; \theta) }{b(\boldsymbol{z}_{<d}; \theta)}
\end{gathered}
\end{equation*}
where $a$ and $b$ are the shift and scale functions with shared parameters $\theta$. The Jacobian matrix is triangular because $\frac{\partial z_i}{\partial \epsilon_j}$ is non-zero only for $j\leq i$, with determinant $\prod b(\boldsymbol{z}_{<d}; \theta)$.

%In these diagrams, circles represent random variables $\epsilon_t$ and $z_t$. Diamonds connecting two random variables represent a scalar affine transformation. Arrows show the dependencies between random variables. Both the generative and the likelihood evaluation process are depicted in this diagram.
A flow diagram of AF is shown in Figure \ref{fig:standard_flows}a.
To sample $\boldsymbol{z}$, we sample each $\epsilon_d$ on the left.
The first $z_1$ is computed through an affine transformation, and then each subsequent $z_d$ is sampled in serial based on $\epsilon_d$ and $\boldsymbol{z}_{<d}$.  To evaluate the density, we simply apply individual scalar affine transformations in parallel, each depending on all previous observed $\boldsymbol{z}_{<d}$, and compute the base density.

%Depicted this way, we can see that the likelihood evaluation process can be performed for all $z_t \rightarrow \epsilon_t$ in parallel because the affine transformations only depend on different $z_t$s, all of which are given before starting the process. Conversely, the generative process is serial for each $\epsilon_t \rightarrow z_t$ because one must generate $z_1$ before all the dependencies for the affine transformation $\epsilon_2 \rightarrow z_2$ are satisfied.

\paragraph{Inverse Autoregressive Flow (IAF)}

Inverse autoregressive flows, proposed in \citet{Kingma2016},
use affine transformations that depend on previous $\boldsymbol{\epsilon}_{<d}$ instead of $\boldsymbol{z}_{<d}$.  The transformation $f_\theta$ for IAF has the form:
\begin{equation*}
\begin{gathered}
\label{eq:iaf}
f_\theta(\boldsymbol{\epsilon})_d = z_d = a(\boldsymbol{\epsilon}_{<d}; \theta) + b(\boldsymbol{\epsilon}_{<d}; \theta) \cdot \epsilon_d \\
f_\theta^{-1}(\boldsymbol{z})_d = \epsilon_d = \frac{z_d - a(\boldsymbol{\epsilon}_{<d}; \theta) }{b(\boldsymbol{\epsilon}_{<d}; \theta)}
\end{gathered}
\end{equation*}

A flow diagram for IAF is shown in Figure \ref{fig:standard_flows}b. For the sampling process all $z_d$ can be computed given $\boldsymbol{\epsilon}$ in parallel; conversely, density evaluation requires computing each $\epsilon_d$ serially.
%In this sense, IAF is the dual of AF as the computational efficiency of training and generation are swapped.
%Importantly, while AF and IAF make up an identical model class.
In practice AF and IAF encode different inductive biases which can hinder the ability of IAF to generalize as well as AF \cite{Oord2017}.

\paragraph{Split Coupling Flow (SCF)}
Split coupling flows, initially proposed in \citet{Dinh2016} and followed up on in \citet{Jul}, utilize "coupling layers" that keep a subset $\mathcal{S} \subset \{1, 2, ..., D\}$ of the random variables unchanged, i.e. $\boldsymbol{z}_{\mathcal{S}} = \boldsymbol{\epsilon}_{\mathcal{S}}$, and use these to condition the transformation for the rest of the random variables $\overline{\mathcal{S}}$.  The transformation $f_\theta$ for SCF and $d \in \overline{\cal S}$ can be written:
\begin{equation*}
\begin{gathered}
\label{eq:realnvp}
%f_\theta(\boldsymbol{\epsilon})_{<S} = \boldsymbol{z}_{<S} = \boldsymbol{\epsilon}_{<S} \\
f_\theta(\boldsymbol{\epsilon})_{d} = z_{d} = a(\boldsymbol{z}_{\mathcal{S}}; \theta)  + b(\boldsymbol{z}_{\mathcal{S}}; \theta) \cdot \epsilon_{d}\\
f_\theta^{-1}(\boldsymbol{z})_{d} = \epsilon_{d} = \frac{z_{d} - a(\boldsymbol{z}_{\mathcal{S}}; \theta) }{b(\boldsymbol{z}_{\mathcal{S}}; \theta)}
\end{gathered}
\end{equation*}
%with inverse
%\begin{equation}
%\begin{gathered}
%\label{eq:realnvp}
%f_\theta^{-1}(\boldsymbol{z})_{<S} = \boldsymbol{\epsilon}_{<S} = \boldsymbol{z}_{<S} \\
%f_\theta^{-1}(\boldsymbol{z})_{\geq S} = \boldsymbol{\epsilon}_{\geq S} = \frac{\boldsymbol{z}_{\geq S} - a(\boldsymbol{z}_{<S}; \theta) }{b(\boldsymbol{z}_{<S}; \theta)}
%\end{gathered}
%\end{equation}
 A flow diagram for SCF is shown in Figure \ref{fig:standard_flows}c, where $S=\{1, 2\}$ for visualization; both sampling and density evaluation are parallel. As SCF is a special case of AF it has a strictly reduced modeling flexibility in exchange for improved computational efficiency \cite{Papamakarios2017}.

 %In the diagram the arrows are shown coming from $z_1$ and $z_2$, but because of the equivalence they could equally have be shown from $\epsilon_1$ and $\epsilon_2$.

\paragraph{Layered Flows}

Each flow encodes an invertible function with a linearly computable Jacobian determinant. Because invertibility is closed under function composition, and the Jacobian determinant factorizes, more flexible distributions can be created by layering flows and changing the ordering of the dependencies at each layer \cite{Salimans2017}. Changing the ordering between layers allows all $z_d$'s or $\epsilon_d$'s to interact with each other, often implemented by reversing or shuffling the ordering of dependencies \cite{Jul}.

Figure \ref{fig:standard_flows}d shows an example with three layers of AF, with reversed ordering between layers. Stacking multiple layers of flow has been shown to significantly increase the modeling flexibility of this class of normalizing flows \cite{Jul,Oord2017}.

A multilayer flow represents a true invertible vector transformation with a dense Jacobian matrix. Forming the building blocks for the discrete flow models, we denote these general multilayers flows $f_{\mathrm{AF}}(\boldsymbol{\epsilon}; \theta)$, $f_{\mathrm{IAF}}(\boldsymbol{\epsilon}; \theta)$, and $f_{\mathrm{SCF}}(\boldsymbol{\epsilon}; \theta)$.

\section{Latent Flows for Discrete Sequences}

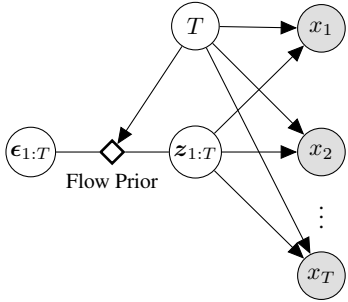
\begin{figure}[t]
\centering
\begin{tikzpicture}[scale=0.8, every node/.style={scale=0.9}]
 % Nodes
 \node[obs]                       (x1)      {$x_1$} ;
 \node[obs, below=of x1]          (x2)      {$x_2$} ;
 %\node[obs, below=of x2]         (x3)      {$x_3$} ;
 \node[below=of x2, yshift=1cm]             (ellipsis)      {$\vdots$} ;
 \node[obs, below=of x2]    (xT)      {$x_T$} ;
 \node[latent, left=of x2]    (z)      {$\boldsymbol{z}_{1:T}$} ; %
 \node[latent, above=of z]   (T)      {$T$} ;
 \node[latent, left=1.5cm of z] (eps) {$\boldsymbol{\epsilon}_{1:T}$};
 \vecinvert[left  = of z]  {eps-z}     {below:Flow Prior} {eps} {z} ; %

 % Edges
 \edge {T} {eps-z, x1, x2, xT} ;
 \edge {z} {x1, x2,  xT} ;

\end{tikzpicture}
\caption{\label{fig:generative_model} Proposed generative model of discrete sequences. The model first samples a sequence length $T$ and then a latent continuous sequence $\boldsymbol{z}_{1:T}$. Each $x_t$ is shown separately to highlight their conditional independence given $\boldsymbol{z}_{1:T}$. Normalizing flow specifics are abstracted by $p(\boldsymbol{z})$ are described in Section~\ref{sec:prior}.}
\end{figure}

Using these building blocks, we aim to develop flexible flow-based models for discrete sequences. Directly working with invertible deterministic mappings in discrete spaces poses significant challenges compared to the continuous setting. Indeed, for one dimensional random variables no such mappings exist besides permutations (details in Supplementary Materials). Instead we explore using a latent-variable model, with a continuous latent sequence modeled through normalizing flows. We begin by describing the full generative process and then focus on the flow-based prior.

\subsection{Generating Discrete Sequences}

Our central process will be a latent-variable model for a discrete
sequence. However, unlike standard discrete autoregressive models, we
aim to lift the main dynamics of the system into continuous space, i.e.
into the prior. In particular, we make the strong assumption that each
discrete symbol is \textit{conditionally independent} given the latent.

Concretely, we model the generation of a discrete sequence
$\boldsymbol{x}_{1:T} =\{x_1, ..., x_T\}$ conditioned on a latent
sequence $\boldsymbol{z}_{1:T}$ made up of continuous random vectors
$\{\boldsymbol{z}_1, ..., \boldsymbol{z}_T\}$ with
$\boldsymbol{z}_t \in \mathbb{R}^H$ and $H$ is a hidden
dimension. Define $p(\boldsymbol{z}_{1:T} | T)$ as our prior
distribution, and generate from the conditional distribution over discrete
observed variables $p(\boldsymbol{x}_{1:T}|\boldsymbol{z}_{1:T},
T)$. The conditional likelihood generates each $x_t$ conditionally
independently:
$p(\boldsymbol{x}_{1:T}|\boldsymbol{z}_{1:T}, T) =\prod_{t=1}^T
p(x_t|\boldsymbol{z}_{1:T}, T)$, where the emission distribution
depends on the dataset.

% Contrast the simple conditionally independent
% factorization of this model to standard time-sequence generative
% models which use an autoregressive likelihood.

To allow for non-autoregressive generation, the length of the sequence
$T$ is explicitly modeled as a latent variable and all parts of the
model are conditioned on it. Length conditioning is elided in the
following discussion (see the Supplementary Materials for
details). The complete graphical model is shown in Figure
\ref{fig:generative_model}.

\begin{figure}[t]
\centering
\includegraphics[width=\linewidth]{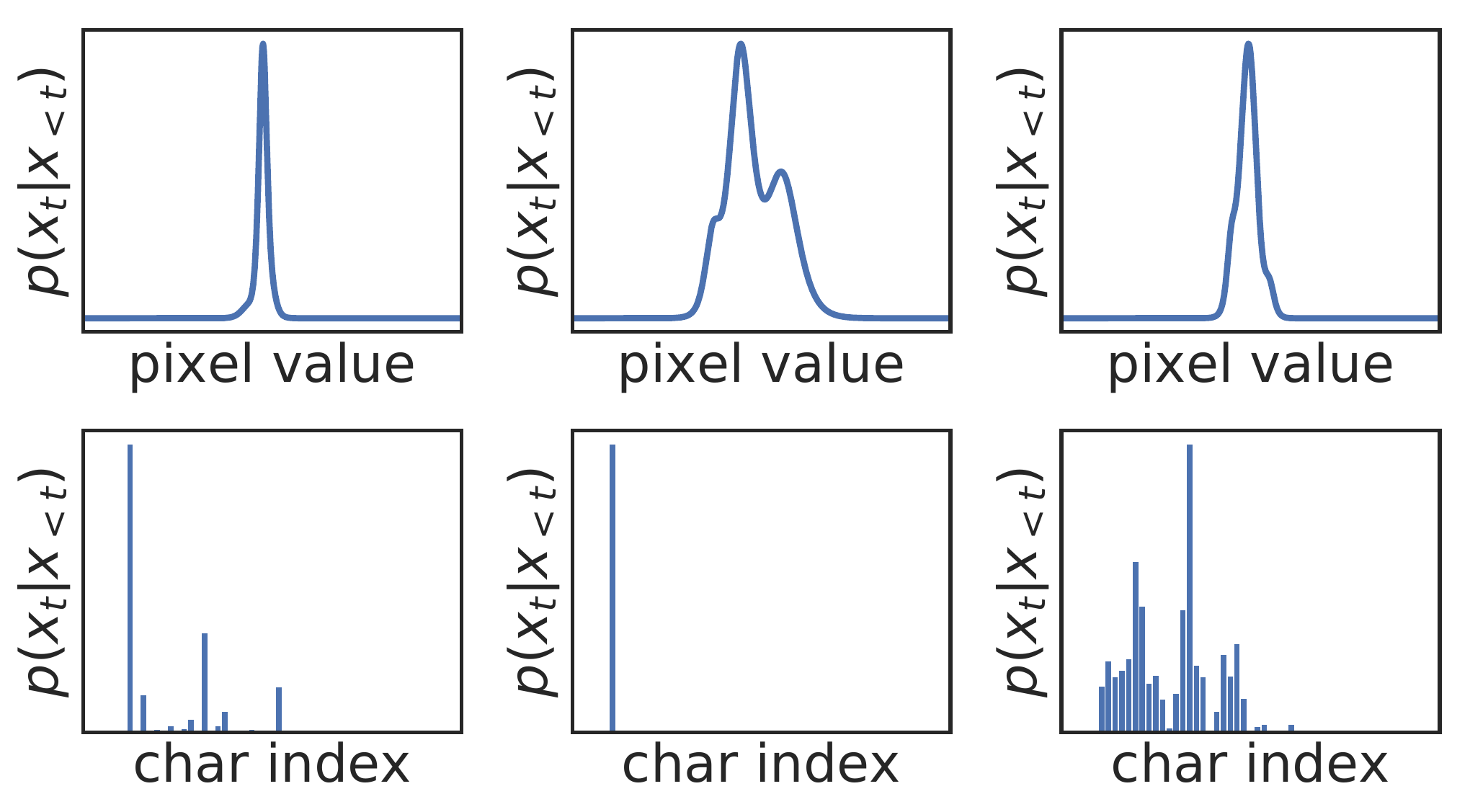}
\caption{\label{fig:multimodality}Example conditional distributions $p(x_t|\boldsymbol{x}_{<t})$ from continuous (PixelCNN++, 10 mixture components, trained on CIFAR-10, top) and discrete (LSTM char-level LM trained on PTB, bottom) autoregressive models.}
\vskip-1.5em
\end{figure}

\subsection{Criteria for Effective Flow Parameterization}

The prior $p(\boldsymbol{z}_{1:T})$ in this process needs to capture the dynamics of the discrete system in a continuous space. Unlike common continuous spaces such as images, in which conditional distributions $p(x_t|\boldsymbol{x}_{<t})$ are often modeled well by unimodal or few-modal distributions, discrete spaces with fixed generation order are highly multimodal.

Figure \ref{fig:multimodality} illustrates this difficulty. First consider the continuous distributions generated by an AF model (PixelCNN++ \cite{Salimans2017}) with 10 mixture components. Despite its flexibility, the resulting distributions have a limited modality indicating that increasing flexibility does not better model the data. Further corroborating this hypothesis, \cite{Salimans2017} report that using more than 5 mixture components does not improve performance.

In contrast, Figure \ref{fig:multimodality}b shows a similar experiment on discrete data. Here the first and third  distributions are highly multimodal (given previous characters there are multiple different possibilities for the next character). Furthermore, the degree of multimodality can vary significantly, as in the second example, requiring models to be able to adjust the number of indicated modes in addition to their locations. In the proposed model, because the conditional likelihood models each $x_t$ as independent, this multimodality at each time step needs to exist almost exclusively in the latent space with each likelihood $p(x_t|\boldsymbol{z})$ being highly constrained in its conditioning.

%For example, one way this could be accomplished is by modeling $p(\boldsymbol{z})$ as an autoregressive mixture model. In that case the mixture components could correspond to individual tokens and the input to the RNN at time step $t$, $z_t$

%The mixture model still contains a discrete distribution in the mixture weights, though, and doesn't have the computational flexibility benefit of flows.

%Compared to generative modeling tasks where conditional densities are largely unimodal (see e.g. \cite{Salimans2017}), if the $\boldsymbol{z}$ space is to completely represent the discrete space of the vocabulary even the conditional densities $p(\boldsymbol{z}_t|\boldsymbol{z}_{<t})$ must be highly multimodal. In addition, this flow needs to model not only the time dimension but also model a vector at each time-step $\boldsymbol{z}_t$. In effect, this requires specifying a flow over $D=TxH$ random variables. In this section we instead consider flow architectures split across both the time and hidden dimension. These differ in their inductive biases as well as sampling and density evaluation process.

%-------------------------------------------------------------------------

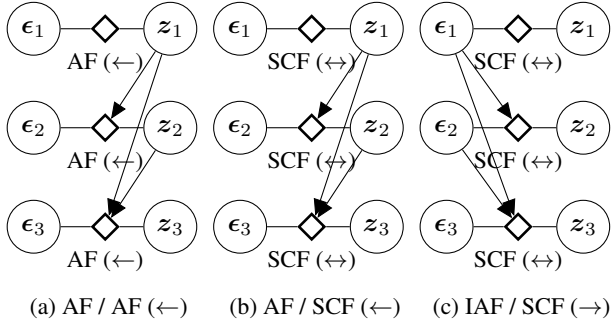
\begin{figure}[t]
\centering
\hfill
\begin{subfigure}[t]{.33\linewidth}
\begin{tikzpicture}
 % Nodes
 \node[latent]                   (eps1)      {$\boldsymbol{\epsilon}_{1}$} ; %
 \node[latent, right=1.1cm of eps1]    (z1)      {$\boldsymbol{z}_{1}$} ;
 \node[latent, below=0.6cm of eps1]    (eps2)      {$\boldsymbol{\epsilon}_{2}$} ; %
 \node[latent, right=1.1cm of eps2]    (z2)      {$\boldsymbol{z}_{2}$} ;
 \node[latent, below=0.6cm of eps2]    (eps3)      {$\boldsymbol{\epsilon}_{3}$} ; %
 \node[latent, right=1.1cm of eps3]    (z3)      {$\boldsymbol{z}_{3}$} ;
% \node[latent, below=0.6cm of eps3]    (eps4)      {$\boldsymbol{\epsilon}{4, :}$} ; %
% \node[latent, right=1.5cm of eps4]    (z4)      {$\boldsymbol{z}_{4, :}$} ;

 % Inverts
 \vecinvertsm[right=of eps1]     {eps-z1}     {below:AF ($\leftarrow$)} {eps1} {z1} ; %
 \vecinvertsm[right=of eps2]     {eps-z2}     {below:AF ($\leftarrow$)} {eps2} {z2} ; %
 \vecinvertsm[right=of eps3]     {eps-z3}     {below:AF ($\leftarrow$)} {eps3} {z3} ; %
% \vecinvert[right=of eps4]     {eps-z4}     {below:$\leftarrow$} {eps4} {z4} ; %

 % Edges
 \path (z1) edge [->, >={triangle 45}] (eps-z2) ;
 \path (z1) edge [->, >={triangle 45}] (eps-z3) ;
% \path (z1) edge [->, >={triangle 45}] (eps-z4) ;
 \path (z2) edge [->, >={triangle 45}] (eps-z3) ;
% \path (z2) edge [->, >={triangle 45}] (eps-z4) ;
% \path (z3) edge [->, >={triangle 45}] (eps-z4) ;

 % Parameters
 %\node[constdot, left=of eps-z1]     (params)   {$\theta$} ;
 %\path (params) edge [->, >={triangle 45}] (eps-z1) ;
 %\path (params) edge [bend left=13, ->, >={triangle 45}] (eps-z2) ;
 %\path (params) edge [bend left=13, ->, >={triangle 45}] (eps-z3) ;
 %\path (params) edge [bend left=13, ->, >={triangle 45}] (eps-z4) ;

 % Ellipsis
 %\node[right=0.4cm of eps-z4]    (ellipsis)      {$\hdots$} ;

\end{tikzpicture}
\caption{AF / AF ($\leftarrow$)}
\end{subfigure}\hfill
\begin{subfigure}[t]{.33\linewidth}
\begin{tikzpicture}
 % Nodes
 \node[latent]                   (eps1)      {$\boldsymbol{\epsilon}_{1}$} ; %
 \node[latent, right=1.1cm of eps1]    (z1)      {$\boldsymbol{z}_{1}$} ;
 \node[latent, below=0.6cm of eps1]    (eps2)      {$\boldsymbol{\epsilon}_{2}$} ; %
 \node[latent, right=1.1cm of eps2]    (z2)      {$\boldsymbol{z}_{2}$} ;
 \node[latent, below=0.6cm of eps2]    (eps3)      {$\boldsymbol{\epsilon}_{3}$} ; %
 \node[latent, right=1.1cm of eps3]    (z3)      {$\boldsymbol{z}_{3}$} ;
 %\node[latent, below=0.6cm of eps3]    (eps4)      {$\boldsymbol{\epsilon}{4}$} ; %
 %\node[latent, right=1.5cm of eps4]    (z4)      {$\boldsymbol{z}_{4}$} ;

 % Inverts
 \vecinvertsm[right=of eps1]  {eps-z1}     {below:SCF ($\leftrightarrow$)} {eps1} {z1} ; %
 \vecinvertsm[right=of eps2]  {eps-z2}     {below:SCF ($\leftrightarrow$)} {eps2} {z2} ; %ove
 \vecinvertsm[right=of eps3]  {eps-z3}     {below:SCF ($\leftrightarrow$)} {eps3} {z3} ; %
% \vecinvert[right=of eps4]  {eps-z4}     {below:$\leftrightarrow$} {eps4} {z4} ; %

 % Edges
 \path (z1) edge [->, >={triangle 45}] (eps-z2) ;
 \path (z1) edge [->, >={triangle 45}] (eps-z3) ;
% \path (z1) edge [->, >={triangle 45}] (eps-z4) ;
 \path (z2) edge [->, >={triangle 45}] (eps-z3) ;
% \path (z2) edge [->, >={triangle 45}] (eps-z4) ;
 %\path (z3) edge [->, >={triangle 45}] (eps-z4) ;

 % Parameters
 %\node[constdot, left=of eps-z1]     (params)   {$\theta$} ;
 %\path (params) edge [->, >={triangle 45}] (eps-z1) ;
 %\path (params) edge [bend left=13, ->, >={triangle 45}] (eps-z2) ;
 %\path (params) edge [bend left=13, ->, >={triangle 45}] (eps-z3) ;
 %\path (params) edge [bend left=13, ->, >={triangle 45}] (eps-z4) ;

 % Ellipsis
 %\node[right=0.4cm of eps-z4]    (ellipsis)      {$\hdots$} ;

\end{tikzpicture}
\caption{AF / SCF ($\leftarrow$)}
\end{subfigure}\hfill
\begin{subfigure}[t]{.33\linewidth}
\begin{tikzpicture}
 % Nodes
 \node[latent]                   (eps1)      {$\boldsymbol{\epsilon}_{1}$} ; %
 \node[latent, right=1.1cm of eps1]    (z1)      {$\boldsymbol{z}_{1}$} ;
 \node[latent, below=0.6cm of eps1]    (eps2)      {$\boldsymbol{\epsilon}_{2}$} ; %
 \node[latent, right=1.1cm of eps2]    (z2)      {$\boldsymbol{z}_{2}$} ;
 \node[latent, below=0.6cm of eps2]    (eps3)      {$\boldsymbol{\epsilon}_{3}$} ; %
 \node[latent, right=1.1cm of eps3]    (z3)      {$\boldsymbol{z}_{3}$} ;
 %\node[latent, below=0.6cm of eps3]    (eps4)      {$\boldsymbol{\epsilon}_{4}$} ; %
 %\node[latent, right=1.5cm of eps4]    (z4)      {$\boldsymbol{z}_{4}$} ;

 % Inverts
 \vecinvertsm[right=of eps1]  {eps-z1}     {below:SCF ($\leftrightarrow$)} {eps1} {z1} ; %
 \vecinvertsm[right=of eps2]  {eps-z2}     {below:SCF ($\leftrightarrow$)} {eps2} {z2} ; %
 \vecinvertsm[right=of eps3]  {eps-z3}     {below:SCF ($\leftrightarrow$)} {eps3} {z3} ; %
% \vecinvert[right=of eps4]  {eps-z4}     {below:$\leftrightarrow$} {eps4} {z4} ; %

 % Edges
 \path (eps1) edge [->, >={triangle 45}] (eps-z2) ;
 \path (eps1) edge [->, >={triangle 45}] (eps-z3) ;
 %\path (eps1) edge [->, >={triangle 45}] (eps-z4) ;
 \path (eps2) edge [->, >={triangle 45}] (eps-z3) ;
 %\path (eps2) edge [->, >={triangle 45}] (eps-z4) ;
 %\path (eps3) edge [->, >={triangle 45}] (eps-z4) ;

 % Parameters
 %\node[constdot, left=of eps-z1]     (params)   {$\theta$} ;
 %\path (params) edge [->, >={triangle 45}] (eps-z1) ;
 %\path (params) edge [bend right=13, ->, >={triangle 45}] (eps-z2) ;
 %\path (params) edge [bend right=13, ->, >={triangle 45}] (eps-z3) ;
 %\path (params) edge [bend right=13, ->, >={triangle 45}] (eps-z4) ;

 % Ellipsis
 %\node[right=0.4cm of eps-z4]    (ellipsis)      {$\hdots$} ;

\end{tikzpicture}
\caption{IAF / SCF ($\rightarrow$)}
\end{subfigure}\hfill
\hfill
\caption{\label{fig:new_flows}Normalizing flows acting on $T$x$H$ random variables proposed in this work. Circles with variables represent random \textit{vectors} of size $H$. Bold diamonds each represent a multilayer AF ($\leftarrow$) or a multilayer SCF ($\leftrightarrow$), as in Figure \ref{fig:standard_flows}d. Arrows to a bold diamond represent additional dependencies to all affine transformations within the indicated AF or SCF.  As above the (arrows) point to the parallel direction, i.e. (a) is parallel in density evaluation whereas (c) is parallel in sampling.
}
\end{figure}

%-------------------------------------------------------------------------

\subsection{Flow Architectures for Sequence Dynamics}
\label{sec:prior}

We consider three flow architectures that describe relations across the time and hidden dimensions that aim to maximize the potential
for multimodal distributions. These differ in their inductive biases
as well as the sampling and density evaluation processes. Note, that
throughout this section $\boldsymbol{z}_t \in \mathbb{R}^H$ represents
a random vector, and so the model is over $D=T\text{x}H$ random
variables. The main concern is the interactions between time $T$
and hidden $H$ dimensions.

%To capture this multimodality, we model a sequence of latent vectors

%the flows studied in this work transform a sequence of vectors. To achieve the high degree of multimodality necessary, we propose the following flows which introduce dependencies between the depth dimension in addition to the time dimension. Like the AF, IAF, and SCF flows, the additional dependency structures determine the computational efficiency of the two directions, likelihood evaluation and generation.

\paragraph{Model 1: AF in time, AF in hidden (AF / AF)}
First consider an autoregressive flow along the time dimension with
each time step applying an autoregressive flow along the hidden
dimension. The transformation function can be written as,
\begin{gather*}
\boldsymbol{z}_{t} = f_{\mathrm{AF}}(\boldsymbol{\epsilon}_{t}; \boldsymbol{z}_{<t}, \theta),\ \ \ \
\boldsymbol{\epsilon}_{t} = f_{\mathrm{AF}}^{-1}(\boldsymbol{z}_{t}; \boldsymbol{z}_{<t}, \theta)
\end{gather*}
where $f_{\mathrm{AF}}(\cdot; \boldsymbol{z}_{<t}, \theta)$ is a layered AF transformation described above with each  constituent affine transformation conditioned on $\boldsymbol{z}_{<t}$ in addition to $\theta$. A proof
that this represents a valid normalizing flow is given in the
Supplementary Materials.

The flow diagram is shown in Figure \ref{fig:new_flows}a. At each time step the AF-in-hidden induces dependencies along the hidden dimension (inside $f_{\mathrm{AF}}$) to create a multimodal distribution. The AF-in-time conditions each subsequent $f_{\mathrm{AF}}$ on the previous latent vectors $\boldsymbol{z}_{<t}$. For density evaluation, $p(\boldsymbol{z}_{1:T})$, both the dependencies within each $f_{AF}$ and the dependencies across time can be computed in parallel. For sampling, each hidden dimension at each time step must be computed in serial.

\begin{figure}[t]
\centering
\hspace*{-0.3cm}\includegraphics[width=1.04\linewidth]{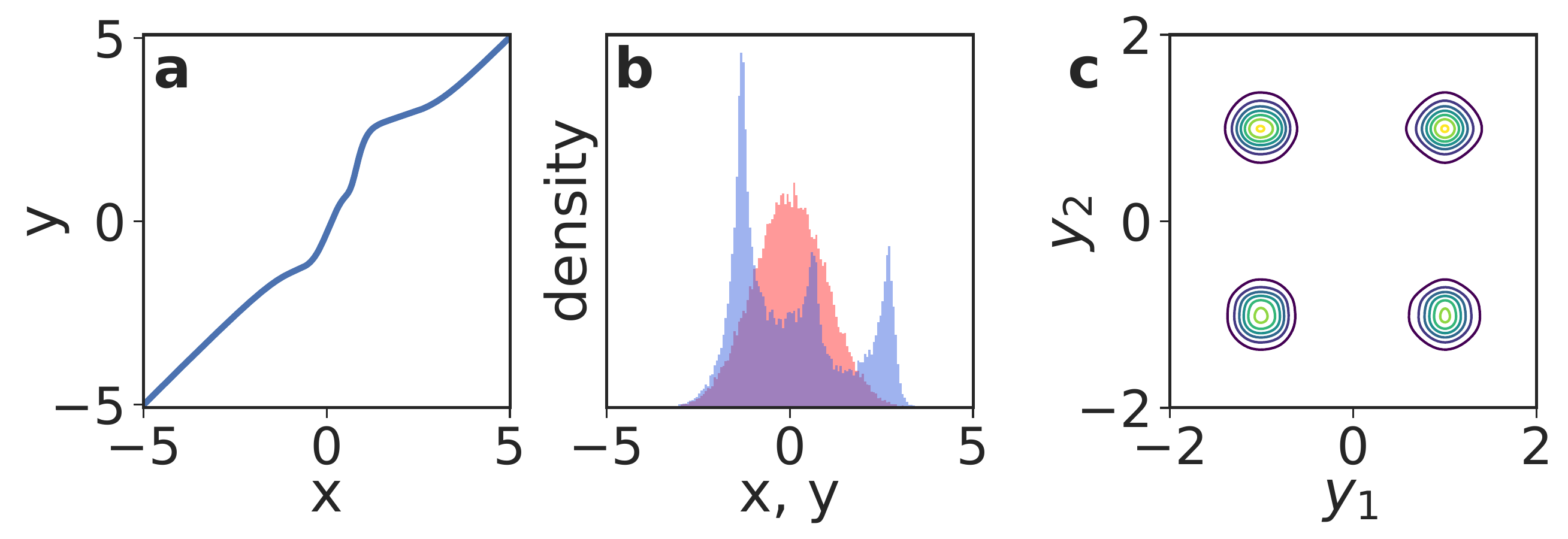}
\caption{Non-Linear Squared (NLSq) flow for multimodal distribution modeling. \textbf{(a, b)} NLSq transformation defined by hand-selecting 4 layers of flow parameters, \textbf{(a)} composed transformation, \textbf{(b)} base density (red), final density (blue). \textbf{(c)} Resulting density for learned 2D transformation via 5 layer AF-like using the NLSq flow from a standard Gaussian to a Gaussian mixture distribution.}
\label{fig:initial_evidence}
\end{figure}

%The flow diagram is shown in Figure \ref{fig:new_flows}a. Here circles represent $D$ dimensional random vectors $\boldsymbol{\epsilon}_t$ and $\boldsymbol{z}_{t}$. Whereas before the diamond represented a scalar affine (invertible) transformation, here the bold diamond labeled $\leftarrow$ represents the vector invertible transformation of a multilayer AF. The explicit arrows to the bold diamonds represent additional dependencies for all the scalar transformations within each bold diamond/AFd.

%For an AFd composed of $J$ layers of flow, we denote the $j$th random vector $\boldsymbol{z}_{t}^{(j)}$, where $\boldsymbol{z}_{t}^{(J)}:=\boldsymbol{z}_t$ and $\boldsymbol{z}_{t}^{(0)}:=\boldsymbol{\epsilon}_t$. Furthermore, we denote the $D$ elements of $\boldsymbol{z}_{t}^{(j)}$ explicitly as $\{z_{t, 1}^{(j)}, ..., z_{t, D}^{(j)}\}$. With this notation, the AFt-AFd flow is defined by the series of $J$ transformation functions:

%AFt-AFd conditions each dimension $z_{t, i}^{(j)}$ on the previous dimensions at the same timestep $\boldsymbol{z}_{t, <i}^{(j)}$ in addition to conditioning on the previous timesteps $\boldsymbol{z}_{<t}^{(J)}$. In \ref{fig:new_flows}a the arrows represent the dependency on previous timesteps $\boldsymbol{z}_{<t}^{(J)}$; the dependencies on previous depth dimensions are hidden within the bold diamond representing a multilayer AF.

\paragraph{Model 2: AF in time, SCF in hidden (AF / SCF)}

Model 1 can be evaluated efficiently, but the serial sampling procedure may be an issue in applications. As an alternative we consider a flow which replaces AF-in-hidden dimension with a layered SCF. The prior is defined by the forward and inverse transformation functions,
\begin{gather*}
\boldsymbol{z}_{t} = f_{\text{SCF}}(\boldsymbol{\epsilon}_{t}; \boldsymbol{z}_{<t}, \theta), \ \ \ \ \boldsymbol{\epsilon}_{t} = f_{\text{SCF}}^{-1}(\boldsymbol{z}_{t}; \boldsymbol{z}_{<t}, \theta)
\end{gather*}
%The AF / SCF model has the same ability to learn statistical relationships over time as AFt-AFd, but is reduced in terms of its flexibility to model a highly multimode distribution.
%In exchange for the reduction in flexibility,

The flow diagram is shown in Figure \ref{fig:new_flows}b.  This model
allows for similar parallel density evaluation as Model 1, however it
is parallel in sampling along the hidden dimension, which can help
efficiency. The downside is that SCF may not be able to induce the
flexible multimodality required for the discrete case.

\paragraph{Model 3: IAF in time, SCF in hidden (IAF / SCF)}

Finally, the autoregressive sampling behavior can be removed  completely.
The final model uses an IAF-in-time to remove this serial dependency in sampling.
The transformation functions are:
\begin{gather*}
\boldsymbol{z}_{t} = f_{\mathrm{SCF}}(\boldsymbol{\epsilon}_{t}; \boldsymbol{\epsilon}_{<t}, \theta), \ \ \ \
\boldsymbol{\epsilon}_{t} = f_{\mathrm{SCF}}^{-1}(\boldsymbol{z}_{t}; \boldsymbol{\epsilon}_{<t}, \theta)
\end{gather*}
The flow diagram is shown in Figure \ref{fig:new_flows}c. For sampling, given $\boldsymbol{\epsilon}_{1:T}$ the time-wise and hidden dependencies can be satisfied in parallel (they all appear on the right side of the forward transformation function). Density evaluation, on the other hand, becomes parallel along hidden and serial in time.\footnote{We also considered an IAF / IAF model; however having fully serial operation in density evaluation makes training prohibitively expensive.}

\iffalse
\begin{figure*}[t]
\centering
\includegraphics[width=1.1\textwidth]{images/initial_evidence.pdf}
\caption{\textbf{(a)}-\textbf{(c)} Transformation defined by hand-selecting 4 layers of flow parameters, demonstrating the ability of the flow to model complicated multimodal distributions: \textbf{(a)} base density, \textbf{(b) }transformation, \textbf{(c)} resulting density. \textbf{(d)} Resulting density for learned 2D transformation via IAF-like training of 5 layers of the NLSq flow from a standard Gaussian to a Gaussian mixture distribution.}
\label{fig:initial_evidence}
\end{figure*}
\fi

\paragraph{Extension: The Non-Linear Squared Flow}

We can add further flexibility to the model by modifying the
core flows. Building on the observations of
\cite{Huang2018}, we propose replacing the affine scalar
transformation with an invertible non-linear squared transformation
(designated NLSq):
\begin{equation*}
f(\epsilon) = z = a + b \epsilon + \frac{c}{1 + \left(d \epsilon + g\right)^2}
\end{equation*}
This transformation has five pseudo-parameters instead of the two for the affine. It reduces to the affine function in the case where $c=0$. When $c\neq 0$, the function effectively adds a perturbation with position controlled by $g$ and scale controlled by $c$ and $d$, which even in 1D can induce multimodality. Under conditions on the scale parameter $c$ the function can be guaranteed to be invertible, and the analytical inverse is the solution to a cubic equation (see Supplementary Materials for details).

Figure \ref{fig:initial_evidence} illustrates the transformation. Figure \ref{fig:initial_evidence}a, b  show an example of four compositions of NLSq functions, and the initial and final density. Whereas the affine transformation would simply scale and shift the Gaussian, the NLSq function induces multimodality. As a final example of the ability of this function to model a multimodal distribution within the flow framework, Figure \ref{fig:initial_evidence}c shows the learned 2D density for a toy dataset consisting of a mixture of four Gaussians. Consistent with \citet{Huang2018}, we find that an AF even with many layers fails to learn to model the same distribution.

% Unless specified, the scalar NLSq function simply replaces the scalar affine transformation represented by a diamond in Figure \ref{fig:standard_flows} in all models, further improving the ability of the proposed flows to model highly multimodal discrete data.

\section{Variational Inference and Training}

To train the model, we need to learn both the simple likelihood and
the prior models.  This requires being able to efficiently perform
posterior inference, i.e. compute the posterior distribution
$p(\boldsymbol{z}_{1:T} | \boldsymbol{x}_{1:T})$, which is
computationally intractable. We instead use the standard approach of
amortized variational inference \cite{Kingma2014} by introducing a
trained inference network,
$q_{\phi}(\boldsymbol{z}_{1:T}|\boldsymbol{x}_{1:T})$. This
distribution $q$ models each $\boldsymbol{z}_t$ as a diagonal Gaussian
with learned mean and variance:
\[q_{\phi}(\boldsymbol{z}_{1:T}|\boldsymbol{x}_{1:T})%=\prod q(\boldsymbol{z}_t|\boldsymbol{x}_{1:T})
=\prod_{t=1}^T \mathcal{N}(\boldsymbol{z}_t|\boldsymbol{\mu}_t(\boldsymbol{x}_{1:T}; \phi), \sigma_t^2(\boldsymbol{x}_{1:T}; \phi) I_H). \]
%While this mean-field factorization results in a weak inference model, \citet{Chen2016a} show that an AF prior is equivalent to an IAF posterior in terms of optimizing the evidence lower bound.  Given that the priors considered in this work are based on AF, a mean-field approximation is sufficient to perform accurate inference.

While this mean-field factorization results in a weak inference model, preliminary experiments indicated that increasing the flexibility of the inference model with e.g. IAF \cite{Kingma2016} did not improve performance.

This inference network is trained jointly with the model to maximize the evidence lower-bound (ELBO),
\[ \log p(\boldsymbol{x}) \geq
\mathbb{E}_{q_{\phi}}\left[\log p(\boldsymbol{x}|\boldsymbol{z})\right] - \text{KL}( q_{\phi}(\boldsymbol{z}|\boldsymbol{x}) \|  p(\boldsymbol{z})) \]
Training proceeds by estimating the expectation with monte-carlo samples and optimizing the lower bound for both the inference network parameters $\phi$ as well as the prior $p(\boldsymbol{z})$ and likelihood $p(\boldsymbol{x}|\boldsymbol{z})$ parameters.

\begin{table}[t]
\centering
\sbox0{\footnotemark}% Store \footnotemark
\sbox1{\footnotemark}% Store \footnotemark
\captionabove{\label{table:results}Character-level language modeling results on PTB. NLL for generative  models is estimated with importance sampling using 50 samples\usebox0, the reconstruction term and KL term refer to the two components of the ELBO. The LSTM from \citet{Cooijmans2017} and AWD-LSTM from \citet{Merity2018} use the standard character-setup which crosses sentence boundaries\usebox1. }
\begin{tabular}{lccc}
\toprule
Model       & Test NLL & Reconst. & KL \\
            & (bpc) & (bpc) & (bpc) \\
\midrule
LSTM      & 1.38 & - & -   \\
%BN-LSTM (\cite{Cooijmans2017})  & 1.32 & - & -   \\
%Zoneout (\cite{Krueger2016})    & 1.27 & - & -   \\
%HM-LSTM (\cite{Chung2016})      & 1.24 & - & -   \\
%HyperNetworks (\cite{Ha2016})   & 1.22 & - & -   \\
%FS-LSTM (\cite{Mujika2017})     & 1.19 & - & -   \\
AWD-LSTM   & 1.18 & - & -   \\
\midrule
LSTM (sentence-wise)  & 1.41 & -    & -      \\
AF-only             &  2.90 & 0.15 & 2.77   \\
\midrule
AF / AF               &  1.42 & 0.10 & 1.37   \\
AF / SCF            &  1.46 & 0.10 & 1.43   \\
IAF / SCF            &  1.63 & 0.21 & 1.55   \\
\bottomrule
\end{tabular}
\end{table}

\section{Methods and Experiments}
\footnotetext[2]{Using more than 50 samples did not further improve results.}

We consider two standard discrete sequence modeling tasks:
character-level language modeling and polyphonic music modeling. For
all experiments, we compute the negative log-likelihood (NLL)
estimated with importance sampling and evaluate on a held-out test set
to evaluate distribution-modeling performance. As a baseline, we use a
LSTM-based language model as in \citet{Press2017}, the standard
discrete autoregressive model. In all flow models a LSTM is used to capture the dependencies across time; the LSTM is the same size as that used in the baseline model. SCF layers are implemented via MADE \cite{Germain2015}. For all flow-based
models, a BiLSTM is used to compute the likelihood model
$p(x_t|\boldsymbol{z})$ and the inference network
$q(\boldsymbol{z}_t|\boldsymbol{x})$. All flow-based models use NLSq
unless otherwise noted.  Optimization and hyperparameter
details are given in the Supplementary Materials.

\subsection{Character-Level Language Modeling}

Character-level language modeling tests the ability of a model to
capture the full distribution of high entropy data with long-term
dependencies. We use the Penn Treebank dataset, with the standard
preprocessing as in \cite{Mikolov2012}. The dataset consists of
approximately 5M characters, with rare words replaced by "$<$unk$>$"
and a character-level vocabulary size of $V=51$.\footnotetext{ Unlike
  previous works on character-level language modeling which consider
  the dataset to be a single continuous string of characters,
  non-autoregressive generation requires the dataset to be split up
  into finite chunks. Following previous text-based VAE works in the
  literature \cite{Bowman2015}, the dataset is split into
  sentences. To avoid extreme outliers, the dataset is limited to
  sentences of length less than 288 tokens, which accounts for 99.3\%
  of the original dataset. Due to these two modifications the absolute
  NLL scores are not precisely comparable between this dataset and the
  one used in previous works, although the difference is small. }

Table \ref{table:results} shows results. The LSTM baseline establishes a "gold standard" representing a model trained directly on the observed discrete sequence with the same $T$ conditioning as the proposed model. A recent state-of-the-art model (AWS-LSTM) with additional orthogonal improvements is also shown for context. In terms of absolute NLL score, AF / AF nearly matches the LSTM baseline, whereas AF / SCF is within 0.05 of the LSTM baseline. These results demonstrate that the combination of AF-in-hidden and the NLSq scalar invertible function induce enough multimodality in the continous distribution to model the discrete data. The AF-only ``unigram'' model removes the relationships across time in the prior model, effectively dropping the time-dynamics.

The IAF / SCF model performs worse than the other models, which reflects the additional challenges associated with non-autoregressive sampling. The same effect is seen with normalizing flow-based generative models for images \cite{Dinh2016,Jul}, where non-autoregressive models have not reached the state-of-the-art performance. Still, compared to the AF-only baseline the autoregressive model clearly learns important dependencies between characters.

Interestingly, in all models the KL term dominates the ELBO, always accounting for over 90\% of the ELBO. This is in stark contrast to previous NLP latent-variable models with strong likelihood models. In these models, the KL term accounts for less than 5\% of the ELBO \cite{Bowman2015,Kim2018,Xu2018}, or less than 30\% of the ELBO when using a specially designed auxiliary loss \cite{Goyal2017}. This indicates that the model 1) is using the latent space to predict each letter, and 2) is rewarded in terms of NLL for accurately encoding the discrete tokens in both the reconstruction term and the KL term.
%Furthermore, the goal of the models are different: in the standard VAE case the goal is to encode the rough meaning of the sentence in the continuous space whereas in the DiscreteFlow case the goal is to encode each token into the continuous space.
% The high KL utilization highlights that the dynamics of the language are being captured in latent space.

%highlights the fact that the VAE optimization procedure can place the vast majority of important information into the latent space when it benefits the model.
\begin{table}[t]
\centering
\captionabove{\label{table:ablation}Ablation experiments. AF / AF is the same result as in Table \ref{table:results}. -NLSq indicates the affine transformation is used instead of NLSq. -AF hidden indicates no dependencies across hidden (an independent vector affine transformation is used instead).}
\begin{tabular}{lccc}
\toprule
Model       & Test NLL & Reconst. & KL \\
            & (bpc) & (bpc) & (bpc) \\
\midrule
AF / AF               & 1.42 & 0.10 & 1.37   \\
\midrule
\ \ - NLSq       & 1.50 & 0.11 & 1.51   \\
\ \ - AF hidden         & 1.57 & 0.14 & 1.57   \\
\ \ - AF hidden and NLSq       & 1.56 & 0.29 & 1.56   \\
\bottomrule
\end{tabular}
\end{table}

Table~\ref{table:ablation} shows model ablations. Without either the NLSq function or the AF-in-hidden dependencies the performance degrades. Once AF-in-hidden is removed, however, further removing NLSq appears to make only a small difference in terms of NLL. These results provide further evidence to our hypothesis that modeling discrete data requires a high degree of multimodality. Furthermore, standard normalizing flows without these additions do not achieve the required flexibility.

\paragraph{Visualizing learned distributions}

%\begin{subfigures}

\begin{figure}[t]
\centering
\includegraphics[width=\linewidth]{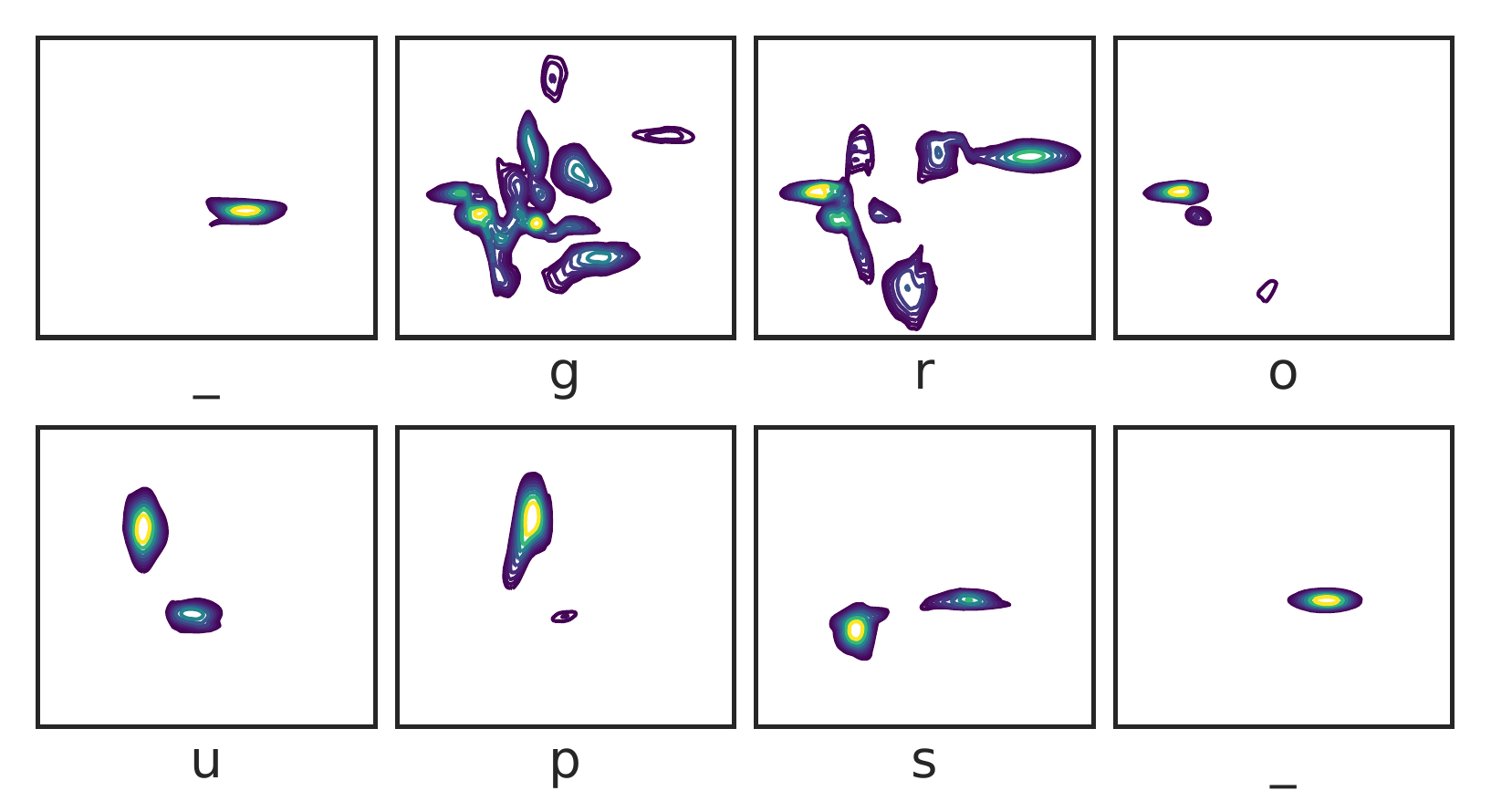}
\caption{\label{fig:priorplots1}Conditional prior densities corresponding to characters in the string `\_groups\_' (\_ indicates a space), from top left to bottom right. Each figure shows $p(\boldsymbol{z}_t | \boldsymbol{z}_{<t})$ for increasing $t$, where $\boldsymbol{z}_{1:T}$ is sampled from $q(\boldsymbol{z}_{1:T}|\boldsymbol{x}_{1:T})$ and $\boldsymbol{x}_{1:T}$ comes from validation.}
\end{figure}

 Figure \ref{fig:priorplots1} shows the prior densities of AF /AF with $H=2$. A continuous sequence of 2-vectors $\boldsymbol{z}$ is sampled from $q(\boldsymbol{z}|\boldsymbol{x})$. The AF / AF model is used to evaluate $p(\boldsymbol{z})$, which gives $p(\boldsymbol{z}_t | \boldsymbol{z}_{<t})$ at every timestep. The figure shows the series of 8 distributions $p(\boldsymbol{z}_t | \boldsymbol{z}_{<t})$ corresponding to the characters ``\_groups\_''. In the first plot we can see that given the previous $\boldsymbol{z}_{<t}$ the prior distribution is unimodal, indicating the model identifies that following the previous word there is only one likely token (a space). At the next timestep, the distribution is highly multimodal, indicating uncertainty of the new word. As the model sees more of the context in the continuous space corresponding to successive characters in the word "groups", the number of modes decreases. In two cases, corresponding to the token following "gro" and the token following "group" the distribution is bimodal, indicating a clear two-way branching decision.
 
%Figure \ref{fig:priorplots}b shows a series of 6 distributions $p(\boldsymbol{z}_t | \boldsymbol{z}_{<t})$ for $t=59$ through $t=64$, corresponding to the string " more" in the same sentence. In this set of plots we see the same pattern as in Figure \ref{fig:priorplots}, where the model is sure of a space at the beginning and end of the word, highly uncertain about the first character of the word, and progressive more certain of the character as $t$ increases. Apparently, while not explicitly trained to do so, the DiscreteFlow model learns a strong one-to-one correlation between individual tokens and the corresponding $\boldsymbol{z}_t$ vectors.

\subsection{Polyphonic Music Modeling}

Next we consider the polyphonic music modeling task \cite{Boulanger-LewandowskiNicolas;BengioYoshua;Vincent2009}. Here each timestep consists of an 88-dimensional binary vector indicating the musical notes played. Unlike character-level language modeling where one token appears at each time step, multiple notes are played simultaneously giving a maximum effective vocabulary size of $2^{88}$. All models are modified so the emission distributions $p(x_t|\boldsymbol{x}_{<t})$ and $p(x_t|\boldsymbol{z})$ are independent Bernoulli distributions instead of Categorical distributions.

Table \ref{table:polyphonic_results} presents the results, split into model classes. RNN/LSTM is the weakest class, capturing the temporal dependencies but treating the 88 notes as independent. RNN-NADE is the strongest class, explicitly modeling the joint distribution of notes in addition to the temporal dependencies. The rest are different latent variable approaches to this problem. They each treat the 88 notes as conditionally independent given a variable-length latent variable.
%By storing useful information in the latent space these models should out-perform the RNN baseline.
All models make different modeling choices and all except DMM include dependencies between observed variables $x_t$.

\begin{table}[t]
\centering
\captionabove{\label{table:polyphonic_results}Polyphonic music likelihood. RNN-NADE is separated to highlight difference in modeling class. Results from our implementation are at the bottom. All numbers are NLL values in nats per note, importance sampling is used to estimate NLL for latent-variable models.  RNN-NADE from \cite{Boulanger-LewandowskiNicolas;BengioYoshua;Vincent2009}, TSBN from \cite{Gan2015}, STORN from \cite{Bayer2014}, NASMC from \cite{Gu2015}, SRNN from \cite{Fraccaro2016}, DMM from \cite{Krishnan2016}.
%All models had  a KL percentage between 49\% and 80\%, exact values are included in the Supplemental Material.
}
\begin{tabular}{lcccc}
\toprule
Model       & Nottingham & Piano & Musedata & JSB \\
\midrule
%RNN      & 4.46       & 8.37        & 8.13       & 8.71          \\
%\midrule
RNN-NADE & 2.31       & 7.05        & 5.6        & 5.19          \\
\midrule
TSBN     & 3.67       & 7.89        & 6.81       & 7.48          \\
STORN    & 2.85       & \textbf{7.13}        &  \textbf{6.16}       & 6.91          \\
NASMC    & 2.72       & 7.61        & 6.89       &\textbf{ 3.99}          \\
SRNN     & 2.94       & 8.2         & 6.28       & 4.74          \\
DMM     & 2.77 &	7.83 &	6.83 &	6.39 \\
\midrule
LSTM  & 3.43 & 7.77   & 7.23 & 8.17     \\
AF / AF   & \textbf{2.39}  & 8.19 & 6.92  & 6.53    \\
AF / SCF  & 2.56  & 8.26  & 6.95  & 6.64    \\
IAF / SCF & 2.54  &  8.25 & 7.06  & 6.59    \\

%AF / AF   & \textbf{2.39} (79\%) & 8.19 (28\%) & 6.92 (56\%) & 6.53 (57\%)   \\
%AF / SCF  & 2.56 (80\%) & 8.26 (33\%) & 6.95 (56\%) & 6.64 (54\%)   \\
%IAF / SCF & 2.54 (65\%) &             &             & 6.59 (49\%)   \\
\bottomrule
\end{tabular}
\end{table}

The AF / AF model outperforms all models on the Nottingham dataset, SRNN on the Piano dataset, and TSBN and STORN on the JSB dataset. The AF / AF model also approaches the RNN-NADE model on the Nottingham dataset. AF / AF performs most poorly on the Piano dataset, which has the longest sequences but only 87 individual sequences. The dataset therefore poorly matches the inductive bias of the discrete flow models, which is designed to ingest whole sequences. The AF / SCF model performs slightly worse than AF / AF on all datasets, which is expected given the loss of modeling power. IAF / SCF performs slightly worse than AF / AF but surprisingly better than AF / SCF on all datasets except Musedata. Given the small amount of training data, IAF / SCF overfits less than AF /SCF, explaining the improved generalization despite being a weaker model.

%This set of results indicates that increasing the modeling flexibility across depth at each time step is generally helpful and more robust to overfitting, whereas too flexible a model across time can lead to overfitting when data is sparse.

Overall, the performance on the polyphonic music datasets demonstrates that the discrete flow model can work at least as well as models which explicitly include connections between the $x_t$s, and that the weakness of the inference model is made up for by the flexibility of the prior.

\subsection{Non-Autoregressive Generation}

\iffalse
\begin{figure}[t]
\centering
\begin{subfigure}[t]{0.49\linewidth}
\includegraphics[width=1\textwidth]{images/timing_analysis_charptb.pdf}
\end{subfigure}
\begin{subfigure}[t]{0.49\linewidth}
\includegraphics[width=1\textwidth]{images/timing_analysis_nottingham.pdf}
\end{subfigure}
\caption{\label{fig:timing}Timing analysis of sentence generation as a function of sequence length, comparing a baseline LSTM model to the IAFt-SCFd model. Character-level language modeling is shown on the left, the Nottingham polyphonic dataset is on the right.}
\end{figure}
\fi

\begin{figure}[t]
\centering
\includegraphics[width=1\linewidth]{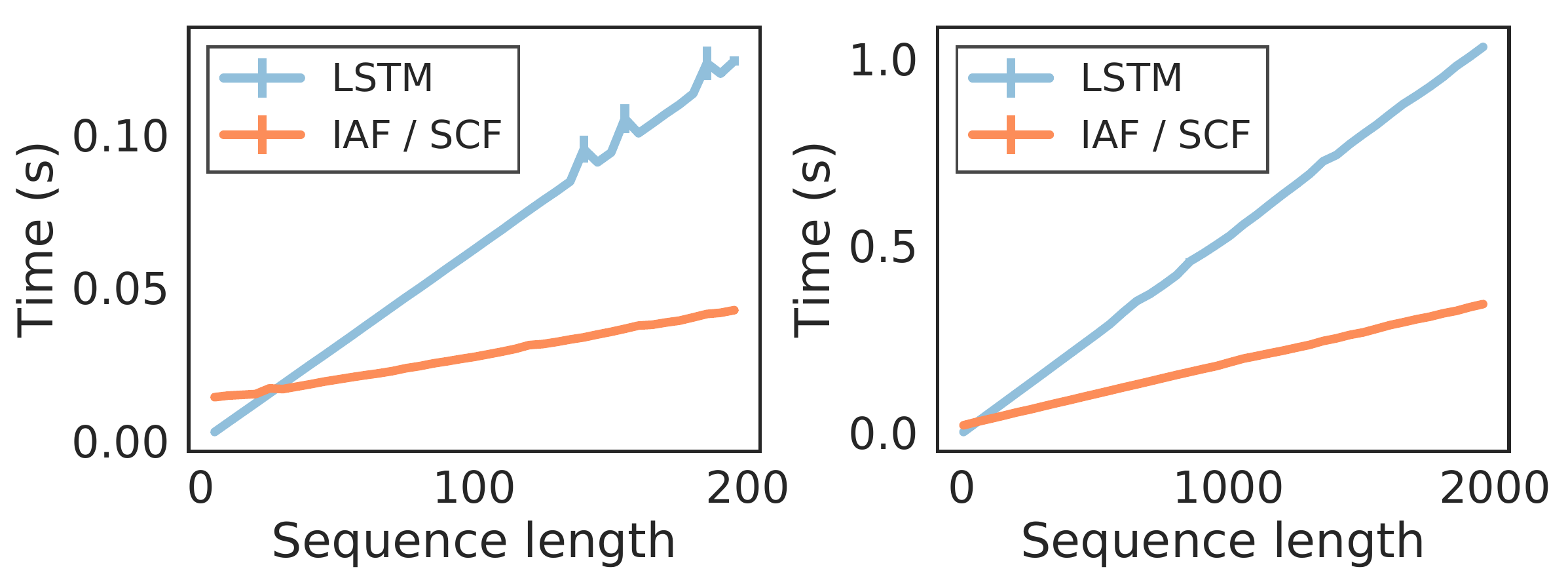}
\caption{\label{fig:timing}Timing analysis of sentence generation as a function of sequence length, comparing a baseline LSTM model to the IAF / SCF model. Character-level language modeling is shown on the left, the Nottingham polyphonic dataset is on the right.}
\end{figure}

While our main goal was to develop a flexible multimodal latent flow model, a secondary goal was to develop a non-autoregressive approach to discrete generation. IAF / SCF best fits this goal. We examine the practical speed of this model compared to discrete autoregressive models.

Figure \ref{fig:timing} shows generation speed for both tasks.  Experiments are run on a single Tesla V100 GPU with a batch size of one, with the  IAF / SCF model using an LSTM to implement time-wise conditioning.
%Here the decoder uses a BiLSTM to translate the $\boldsymbol{z}$ vectors to the $\boldsymbol{x}$ output.
Compared to the baseline LSTM model, the speedup  comes from the fact that in the IAF formulation all of the inputs $\boldsymbol{\epsilon}$ are available to the LSTM in parallel and therefore cuDNN can parallelize parts of the computation.

%Compare this to the baseline LSTM model, which must compute e.g. $x_2$ before feeding $x_2$ back in to the LSTM to compute $x_3$.

Figure \ref{fig:timing} shows that for very short sequences the overhead of the proposed model makes generation slower than the baseline LSTM, whereas after that point the IAF / SCF is faster than the LSTM. This experiment was run with a batch size of 1, for small batch sizes the trend holds while for large batch sizes the additional parallelization afforded by having access to all LSTM inputs becomes less important.

%Compared to models which require an autoregressive factorization in the data space, our decomposition allows for non-autoregressive generation, a strong latent representation of the data, and a nearly fully differentiable generative process (up to the emission). 

\section{Conclusion}

This work proposes a latent-variable model for discrete sequences that learns a highly multimodal normalizing flow-based continuous distribution. We show that two flows, AF / AF and AF / SCF, succeed in learning rich multimodal distributions. Furthermore, we show that IAF / SCF, while slightly less accurate, is an efficient approach for non-autoregressive generation.

The combination of continuous prior and inputless decoder imparts interesting properties that are worth further study in their own right. Future work should investigate applications of the strong continuous representations learned by the flow and the nearly fully differentiable generative process. The latter can be especially useful for GANs, which have traditionally encountered challenges in their application to discrete data such as text.

While LSTM-based unconditional models have been considered here as a proof-of-concept, future work can consider adapting the models for conditional language modeling or moving to alternate architectures such as those based on self-attention. Finally, we hope this work encourages further exploration of the interplay between and relative merits of discrete and continuous representations.

\section*{Acknowledgements}
Thanks to Yoon Kim for useful feedback and discussion.  The authors acknowledge the support of NSF 1704834, CAREER 1845664, Intel, and Google. 

\bibliography{main}
\bibliographystyle{icml2019}

\twocolumn[
\icmltitle{Supplementary Materials}
]

\appendix

\section{Invertible discrete mappings}

It is interesting to consider how one might directly apply flows to discrete sequences. We begin with the discrete change of variables formula: for $X \in \Omega_x^D$, $Y \in \Omega_y^D$ (with $\Omega_x,\Omega_y$ finite), base density $p_X(\boldsymbol{x})$, and deterministic function $\boldsymbol{y}=f(\boldsymbol{x})$,

\begin{align*}
    p_Y(\boldsymbol{y}) = \sum_{\boldsymbol{x} \in f^{-1}(\boldsymbol{y})}p_X(\boldsymbol{x})
\end{align*}

If $f$ is invertible, as is required for the flow framework, this reduces to

\begin{align*}
    p_Y(\boldsymbol{y}) = p_X(f^{-1}(\boldsymbol{y}))
\end{align*}

First, examine the simplest case where $D=1$ and $X \in \Omega_x$, $Y \in \Omega_y$ i.e. $x,y$ are just single elements of a set. In this case, invertible functions can only be found if $|\Omega_x|=|\Omega_y|$, so without loss of generality we can rename elements such that $\Omega_x=\Omega_y=\Omega$. Thus we are interested in invertible functions $f:\Omega \rightarrow \Omega$. By definition, a permutation of $\Omega$ is any invertible mapping from $\Omega$ to itself \cite{Nering1970}. We conclude that when $D=1$, the only possible invertible functions are permutations. As permutations do not permit a parameterized changing of densities, a normalizing flow cannot be used in the 1D case to define a useful distribution.

This is not just a theoretical result, consider the following example: we are interested in learning the distribution of the first word in a set of documents. In this case $\Omega$ would be the vocabulary of possible words, $y$ represents a word, and we would like to use a discrete flow to model $p(y)$. We pick an uninformative base density such as the uniform distribution. According to the result above, a flow cannot learn the distribution $p(y)$, whereas simply counting would model the distribution well.

Even if we were to choose a different distribution, say a geometric distribution in some order, the flow could at best find a permutation of the geometric probabilities that best matches the true distribution. Clearly this is an undesirable optimum.

In the more general case where $D>1$, non-permutation invertible mappings certainly exist. A common example is the XOR function. Therefore, it is in principle possible to use flows to model data when $D>1$, and future work should investigate the limits of this approach. Given that the 1D case fails, however, it will be important to understand how this failure relates to the higher dimensional cases. Is this simply an unfortunate edge case? Or is it indicative of a larger limitation?

\section{Proposed flow validity}

A transformation function $f: \mathcal{R}^D \rightarrow \mathcal{R}^D$ represents a valid normalizing flow if $f$ is invertible. A transformation function $f$ represents a \textit{useful} normalizing flow is the Jacobian of $f$ can be computed with linear complexity in dimension of the data. We show that the three proposed flows in this work have both of these properties.

First consider the AF / AF flow, defined by transformation function $f_\theta$:
\begin{gather*}
\boldsymbol{z}_{t} = f_{\mathrm{AF}}(\boldsymbol{\epsilon}_{t}; \boldsymbol{z}_{<t}, \theta)
\end{gather*}
To prove the mapping is invertible it suffices to find the inverse:
\begin{gather*}
\boldsymbol{\epsilon}_{t} = f_{\mathrm{AF}}^{-1}(\boldsymbol{z}_{t}; \boldsymbol{z}_{<t}, \theta)
\end{gather*}
$f_{AF}$ is a normalizing flow and therefore an invertible function. Each $\boldsymbol{\epsilon}_t$ can thus be calculated from $\boldsymbol{z}_{1:T}$ giving $\boldsymbol{\epsilon}_{1:T}=f_{\theta}^{-1}(\boldsymbol{z}_{1:T})$. 

For the latent flows considered in the main text $\boldsymbol{z} \in \mathcal{R}^{T\text{x}H}$. Here we equivalently view $\boldsymbol{z}$ as a large $D=T\cdot H$ vector. We write $\boldsymbol{z}=\boldsymbol{z}_{1:T}=\{\boldsymbol{z}_1, ..., \boldsymbol{z}_T\}=\{z_{1, 1}, ..., z_{1, H}, z_{2, 1} ..., z_{2, H}, ..., z_{T, 1}, ..., z_{T, H}\}$. In this case the Jacobian matrix $\frac{\partial \boldsymbol{z}}{\partial \boldsymbol{\epsilon}}$ can be written as a block matrix

\begin{equation*}
\begin{bmatrix}
    \frac{\partial \boldsymbol{z}_1}{\partial \boldsymbol{\epsilon}_1} & \dots &   \frac{\partial \boldsymbol{z}_1}{\partial \boldsymbol{\epsilon}_T} \\
    \vdots & \ddots & \vdots \\
   \frac{\partial \boldsymbol{z}_T}{\partial \boldsymbol{\epsilon}_1} & \dots &   \frac{\partial \boldsymbol{z}_T}{\partial \boldsymbol{\epsilon}_T}
\end{bmatrix}
\end{equation*}
where each block $ \frac{\partial \boldsymbol{z}_t}{\partial \boldsymbol{\epsilon}_s}$ is a $H\text{x}H$ Jacobian matrix.

For the AF / AF flow $ \frac{\partial \boldsymbol{z}_t}{\partial \boldsymbol{\epsilon}_s}=\boldsymbol{0}; s>t$ because $\boldsymbol{z}_t$ depends only on $\boldsymbol{\epsilon}_t$ and $\boldsymbol{z}_{<t}$, which itself only depends on $\boldsymbol{\epsilon}_{<t}$. Therefore the Jacobian matrix is block triangular with determinant
\begin{gather*}
\left| \frac{\partial \boldsymbol{z}}{\partial \boldsymbol{\epsilon}} \right|=\prod_{t=1}^T \left| \frac{\partial \boldsymbol{z}_t}{\partial \boldsymbol{\epsilon}_t} \right| =\prod_{t=1}^T \left| \frac{\partial f_{AF}}{\partial \boldsymbol{\epsilon}_t} \right| 
\end{gather*}
Thus, the Jacobian determinant is simply the product of the Jacobian determinants of the AF-in-hidden transformations at each time step. \cite{Papamakarios2017} show that the Jacobian determinant is linear in $H$ for AF, thus the overall complexity for the determinant calculation of AF / AF is $\mathcal{O}(TH)=\mathcal{O}(D)$.

The proof holds when $f_{AF}$ is replaced with $f_{SCF}$, as \cite{Dinh2016} show that the Jacobian of $f_{SCF}$ can be computed with linear complexity. This concludes the proof that AF / AF and AF / SCF are valid normalizing flows with Jacobian determinant calculations linear in the data dimension.

For IAF / SCF the transformation function pair is:
\begin{gather*}
\boldsymbol{z}_{t} = f_{\mathrm{SCF}}(\boldsymbol{\epsilon}_{t}; \boldsymbol{\epsilon}_{<t}, \theta), \ \ \ \
\boldsymbol{\epsilon}_{t} = f_{\mathrm{SCF}}^{-1}(\boldsymbol{z}_{t}; \boldsymbol{\epsilon}_{<t}, \theta)
\end{gather*}
This is invertible because an inverse function is found. $ \frac{\partial \boldsymbol{z}_t}{\partial \boldsymbol{\epsilon}_s}=\boldsymbol{0}; s>t$ because $\boldsymbol{z}_t$ depends only on $\boldsymbol{\epsilon}_t$ and $\boldsymbol{\epsilon}_{<t}$. The Jacobian matrix is thus block triangular with determinant $\prod_{t=1}^T \left| \frac{\partial f_{SCF}}{\partial \boldsymbol{\epsilon}_t} \right| $. The same argument as for AF / AF gives a Jacobian determinant complexity of $\mathcal{O}(TH)=\mathcal{O}(D)$.

\section{NLSq invertibility}

The NLSq function is
\begin{equation}
\label{eq:nlsq}
f(\epsilon) = z = a + b \epsilon + \frac{c}{1 + \left(d \epsilon + g\right)^2}
\end{equation}
In the following discussion we assume $b>0, d>0$ A real scalar function is invertible if its derivative is positive everywhere. 

\begin{equation*}
f'(\epsilon) = b - \frac{2cd(d\epsilon+g)}{(1 + \left(d \epsilon + g\right)^2)^2}
\end{equation*}
Taking another derivative and setting it equal to 0 gives the critical points $\epsilon^*=(g \pm \sqrt{1/3})/d$. The distinction between maximum and minimum depends on the sign of $c$. In either case, the minimum slope is
\begin{equation*}
f'(\epsilon^*) = b - \frac{9}{8\sqrt{3}}|c|d
\end{equation*}
Thus invertibility is guaranteed if $b >  \frac{9}{8\sqrt{3}}|c|d$. In our implementation $a=a$, $g=g$, $b=e^{b'}$, $d=e^{d'}$, and $c=\frac{8\sqrt{3}}{9d}b \alpha \cdot \text{tanh}(c')$, where $a,b',c',d',g$ are unrestricted and output from the model, and $0<\alpha<1$ is a constant included for stability. We found $\alpha = 0.95$ allows significant freedom of the perturbation while disallowing "barely invertible" functions.

The inverse of the NLSq function is analytically computable, which is important for efficient generation. Solving for $\epsilon$ in Eq. \ref{eq:nlsq} gives the cubic equation

\begin{gather*}
-bd^2\epsilon^3 + ((z-a)d^2 - 2dgb)\epsilon^2 \\
+ (2dg(z-a)-b(g^2+1))\epsilon \\
+ ((z-a)(g^2+1)-c) = 0
\end{gather*}
Under the invertibility condition above this is guaranteed to have one real root which can be found analytically \cite{G.C.Holmes2002}.

In practice, because the forward direction as written (applying $f(\epsilon)$) requires fewer operations it is used for the reverse function $f^{-1}(z)$, and the solution to the cubic equation is used for the forward function $f(\epsilon)$.

\section{Variable length input}

When working with non-autoregressive models we need to additionally deal with the variable length nature of the observed sequences. Unlike autoregressive models, which can emit an end-of-sentence token, non-autoregressive models require the length to be sampled initially. Given a sequence of length $T$ we can write

\begin{equation*}
p(\boldsymbol{x}) = \int p(\boldsymbol{x}|T')p(T')\text{d}T' = p(\boldsymbol{x}|T)p(T)
\end{equation*}
where the second equality comes from the fact that $p(\boldsymbol{x}|T')=0$ for $T'\neq T$. For unconditional sequence modeling we can use the empirical likelihood for $p(T)$, and then condition all parts of the model itself on $T$. In this work we implement the conditioning as a two one-hot vectors at every timestep $t$, indicating the distance from the beginning and end of the sequence. Compared to other popular position encodings in the literature, such as the one commonly used in the Transformer \cite{Vaswani2017}, this primarily encodes the absolute length $T$ instead of the relative position between tokens needed in a self-attention based architecture.

The generative process becomes:
\begin{gather*}
T \sim p(T) \\
\boldsymbol{\epsilon} \sim p_\epsilon(\boldsymbol{\epsilon}) \\
\boldsymbol{z} = f_\theta(\boldsymbol{\epsilon};T) \\
\boldsymbol{x} \sim p(\boldsymbol{x}|\boldsymbol{z}, T)
\end{gather*}
\section{Implementation and optimization details}

During optimization, the expectation in the ELBO is approximated with 10 samples. 5 layers of AF-in-hidden or SCF-in-hidden flow are used for the AF / AF and AF / SCF models and 3 layers are used for the IAF / SCF models, for character-level language modeling. 5 layers of SCF-in-hidden are used for all models on the polyphonic datasets. The base density is a standard Gaussian. Adam is used as the optimizer with a learning rate of 1e-3 and a gradient clipping cutoff of 0.25. Dropout is used to regularize the baseline model and the LSTM in the prior of the AF / AF and AF / SCF models. All LSTMs are two layers deep, and all embedding and hidden layers are made up of 500 units. Weight tying between the input embedding of the encoder and output embedding of the decoder is employed.

A latent size of $D=50$ for each random vector $\boldsymbol{z}_t$ and $\boldsymbol{\epsilon}_t$ is used. During preliminary experiments we found that for character-level language modeling the results were nearly identical for $D=5-80$. 

Many recent works have found that it is necessary to bias the variational optimization to prevent posterior collapse, most commonly by using KL annealing or modifying the objective \cite{Bowman2015,Kingma2016,Chen2016a}, without which it is easy for the model to obtain strong performance by simply ignoring the latent code. In our case we similarly find that KL annealing is essential to learn a strong mapping. We hypothesize that while the decoder is extremely weak, the prior itself is powerful and thus the generative model overall is still powerful enough to require such a bias.

Specifically, for the language modeling task we use KL annealing with an initial period of 0 weight on the KL term for 4 epochs followed by a linear increase to the full ELBO across 10 epochs. This schedule allows the models to first encode the vocabulary in the continuous space with 0 reconstruction loss and then learn the statistical dependencies between tokens. For the polyphonic datasets we extend this to 0 weight for 20 epochs followed by a linear increase over 15 epochs, due to the reduced dataset size.

\end{document}